\documentclass[lettersize,journal]{IEEEtran}
\usepackage{amsmath,amsfonts}
\usepackage{algorithmicx}
\usepackage{algorithm}
\usepackage{algpseudocode}
\usepackage{array}
\usepackage{textcomp}
\usepackage{stfloats}
\usepackage{url}
\usepackage{verbatim}
\usepackage{graphicx}
\usepackage{cite}
\usepackage{color}
\usepackage{graphics,caption} 
\graphicspath{ {figure/} }
\usepackage{epsfig} 
\usepackage{times} 
\usepackage{amssymb} 
\usepackage{booktabs}
\usepackage{threeparttable}
\usepackage{multirow}
\if CLASSOPTIONcompsoc
\usepackage[caption=false, font=normalsize, labelfont=sf, textfont=sf]{subfig}
\else
\usepackage[caption=false, font=footnotesize]{subfig}

\usepackage[colorlinks,linkcolor=blue]{hyperref}

\usepackage{float}
\usepackage{stfloats}
\pdfoutput=1

\begin{document}

\title{VLN-Game: Vision-Language Equilibrium Search for Zero-Shot Semantic Navigation}

%

 \author{
     Bangguo Yu$^{1, 2}$, Yuzhen Liu$^1$$^{\dagger}$, Lei Han$^1$, Hamidreza Kasaei$^2$$^{\dagger}$, Tingguang Li$^1$$^{\dagger}$, and Ming Cao$^2$ \\
     \vspace{1em}
     $^1$Tencent Robotics X  \quad\quad    $^2$University of Groningen
 \thanks{$^{\dagger}$ corresponding author }
 \thanks{Part of this work was conducted during the first author’s internship at Tencent Robotics X.}
    
 }



\maketitle


\begin{abstract}
    Following human instructions to explore and search for a specified target in an unfamiliar environment is a crucial skill for mobile service robots. Most of the previous works on object goal navigation have typically focused on a single input modality as the target, which may lead to limited consideration of language descriptions containing detailed attributes and spatial relationships. 
    To address this limitation, we propose VLN-Game, a novel zero-shot framework for visual target navigation that can process object names and descriptive language targets effectively. To be more precise, our approach constructs a 3D object-centric spatial map by integrating pre-trained visual-language features with a 3D reconstruction of the physical environment. Then, the framework identifies the most promising areas to explore in search of potential target candidates. A game-theoretic vision-language model is employed to determine which target best matches the given language description.
    Experiments conducted on the Habitat-Matterport 3D (HM3D) dataset demonstrate that the proposed framework achieves state-of-the-art performance in both object goal navigation and language-based navigation tasks. Moreover, we show that VLN-Game can be easily deployed on real-world robots. The success of VLN-Game highlights the promising potential of using game-theoretic methods with compact vision-language models to advance decision-making capabilities in robotic systems.
    The supplementary video and code can be accessed via the following link: \href{https://sites.google.com/view/vln-game}{https://sites.google.com/view/vln-game}.
\end{abstract}


\section{Introduction}
\begin{figure}[htbp]
    \centering
    \includegraphics[scale=0.43]{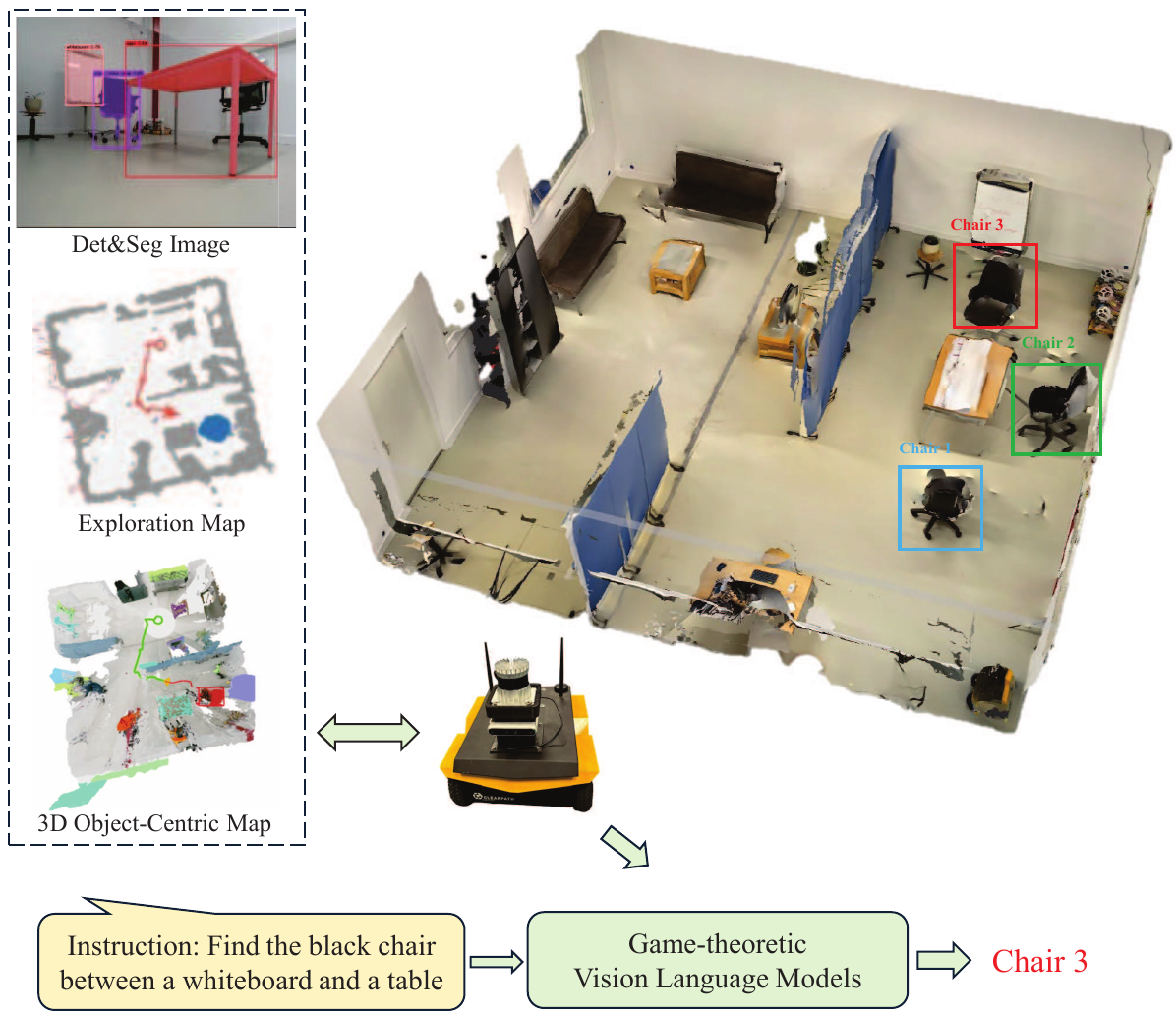}
    \caption{Visual target navigation example. The robot identifies the chair that best matches the description of the instruction in such a complex and unstructured environment using game-theoretic vision language models.}
    \label{fig:demo}
    \vspace{-0.3cm}
\end{figure}

Exploring and grounding a special target in an unknown and complex environment is crucial for intelligent robots in embodied tasks, which, when mimicking human behaviors, requires robots to possess semantic reasoning abilities. For instance, imagine a robot walking into your room, and you say: ``Hey, robot, can you check if my phone is on the black chair between the TV and desk in the living room? After finding it, bring it to me.'' In such a high-level, descriptive task, the first step of the robot is to locate the specified chair. 
Although significant progress has been made in object goal navigation, enabling robots to efficiently find a target instance based on visual input, grounding a language-described target with specific attributes and spatial relationships remains challenging. In this description ``the black chair between the TV and the desk", the attribute ``black" and the spatial relationship ``between the TV and desk" are essential for accurately grounding the target ``chair."
In this work, we focus on the visual target navigation task, which requires robots to explore an unknown 3D environment efficiently in order to locate the object, guided by either the target's category or a language-based description.

Visual target navigation has gained significant research attention, driven by advancements in fast simulation \cite{aihabitat} \cite{Gibson} \cite{ai2thor}, large-scale photo-realistic 3D scene datasets \cite{HM3D}\cite{Gibson}\cite{MP3D}, scene representation \cite{nerf}\cite{VLMaps}\cite{3dgs}, and vision-language models \cite{clip}\cite{llava}\cite{openai2023gpt4}. Depending on the various input types, the task can be divided into object goal and language goal navigation.
Previous studies have primarily focused on object goals, typically using end-to-end \cite{Zhu2017}\cite{Wijmans2019}\cite{Ramrakhya2022} or module-based\cite{SemExp}\cite{Chaplot2020}\cite{poni} frameworks to enhance navigation performance through reinforcement learning and supervised learning techniques.
With the development of large language models (LLMs) and vision-language models (VLMs), a branch of work has explored leveraging these models' scene understanding and reasoning abilities for navigation tasks \cite{Yu2023a}\cite{goat}\cite{VLMaps}\cite{vlfm}. 
While much work has concentrated on object goals, fewer studies have addressed language-based descriptive goals that incorporate attributes and spatial relationships, which is the focus of this work. Given the complexity of language descriptions, prior research has attempted to utilize LLMs or VLMs to interpret scene representations and reason about spatial relationships \cite{LLMGrounder}\cite{navgpt}\cite{discussnav}\cite{instructnav}\cite{khanna2024goatbench}, subsequently making decisions based on the insights provided by these large models.
While these approaches show potential, there is still significant room for improvement, as there remains a great challenge for large models to understand the 3D environments. In addition, the responses from the large models are always noisy and conflict-prone since the large models are far from perfectly reliable in complex tasks \cite{hallucinate}\cite{mckenzie2023inverse}\cite{consensusgame}.

In this paper, we focus on efficient robot navigation and searching for an object with a category's name or language description in an unknown environment based on visual observation. Our proposed approach integrates game-theoretic vision-language models into a novel zero-shot visual target navigation framework, called VLN-Game, to efficiently explore and identify complex targets. Specifically, an object-centric 3D map is constructed during exploration, combining pre-trained CLIP \cite{clip} image features with a 3D point cloud of the environment. The navigation frontier is then determined based on the CLIP similarity map between the target's description and the 3D map. To accurately identify the target that best matches the user's description, top-view and first-person view images of each candidate target are fed into the vision-language models to understand the scenes and make the final decision. We operationalize these high-level objectives using a game-theoretic framework in which equilibrium strategies are calculated to enhance the coherence of the vision-language models and reduce hallucinations during navigation.
An illustrative example of visual target navigation for finding a standing person between a whiteboard and a bar stool is shown in Fig. \ref{fig:demo}. After scanning the environment, the robot detects several people near the whiteboard or chair and needs to determine which one matches the given description. Based on multi-view observations of each candidate and the provided instructions, the game-theoretic vision-language model selects the most relevant target (person 4).
We extensively evaluate our method against state-of-the-art approaches on object-goal and language-goal datasets within the HM3D \cite{HM3D} environment, using the AI-Habitat \cite{aihabitat} platform. Experimental results demonstrate that our method outperforms existing zero-shot object-goal navigation techniques and shows significant advantages in language-based navigation tasks. These findings underscore the potential of game-theoretic tools to improve coherence in vision-language models, leading to enhanced accuracy in visual target navigation tasks. Furthermore, we also showcase the application of our method in real-world settings with various navigation instructions.

In summary, this article makes the following key contributions.
\begin{itemize}
    \item We introduce a novel game-theoretic vision-and-language navigation framework capable of exploring unknown environments and identifying complex targets using either the object's category or language description. 
    \item A new frontier-based semantic exploration policy that multiple similarity value maps are proposed for selecting the most relevant frontier based on the target and the current observation, which achieves the best success rate compared with current state-of-the-art methods in the zero-shot object-goal navigation task.
    \item We propose a strategy of integrating multi-view images of each candidate object from the 3D map into vision-language models to enhance the model's understanding of the 3D environment and the ability to identify complex descriptive targets.
    \item We make the first effort to integrate the equilibrium-seeking strategy in the decision-making of the vision-language model during the navigation, and the application with a tiny model outperforms larger models, demonstrating the effectiveness of game-theoretic tools for formalizing and improving coherence in visual target navigation tasks.
    \item We validate our method on a real robot equipped with an RGB-D camera in diverse and unknown office environments, achieving a real-time lifelong navigation system, which demonstrates the applicability and effectiveness.
\end{itemize}

The rest of this article is organized as follows.  Section II reviews related works for visual target navigation. Section III details the technical aspects of the method. Section IV evaluates and compares the proposed method with previous methods. Finally, Section V presents the conclusions.


\section{Related Work}
\label{sec:citations}
Visual target navigation is one of the fundamental tasks for intelligent robots, inspired by human abilities for semantic reasoning and object localization. In this section, we review related work, starting from classical navigation approaches to visual target navigation using different input modalities, as well as the application of game-theoretic methods in large models.

\subsection{Classical Semantic Navigation}	
Classical approaches to visual target searching aim to construct semantic maps of the environment to capture spatial awareness and historical information using visual observations. These maps are then used to determine a path from the starting point to the desired destination. Various types of maps have been proposed, including occupancy maps \cite{Hess}\cite{orb-slam}\cite{fastlio}, topological graphs \cite{Armeni}\cite{Rosinol2020}, semantic maps\cite{kimera}\cite{Grinvald2019}, and implicit maps \cite{nerf}\cite{VLMaps}\cite{3dgs}. Once the path is planned based on the map, target navigation essentially becomes an obstacle avoidance problem.
However, the effectiveness of these approaches is highly dependent on the accuracy of the environment map. This limitation poses a challenge in unexplored or dynamic environments, where maps cannot be constructed perfectly in advance. To address this issue, frontier-based exploration \cite{frontier} is commonly employed to incrementally explore the environment and build the map.
Recent work has introduced a range of visual navigation tasks, categorized by how the goal is specified: point-goal navigation \cite{Wijmans2019}, image-goal navigation \cite{Zhu2017}, object-goal navigation \cite{SemExp}, and language-goal navigation \cite{khanna2024goatbench}\cite{vln}. This paper specifically focuses on visual target navigation guided by object categories and language descriptions.

\subsection{Visual Target Navigation With Object Category}
Visual target navigation using object categories is a classical task in the field of Embodied AI, where an agent must locate a specific object instance in an unfamiliar environment based on visual observations. Learning-based methods, particularly those that optimize navigation policies through end-to-end or modular approaches, have attracted significant attention and made considerable progress in this area. Most of these methods formulate the task as a reinforcement learning (RL) problem, developing various representations of scene features to improve navigation.
For the end-to-end models, the first framework \cite{Zhu2017} introduced a framework that used a pre-trained ResNet \cite{resnet} to encode both the input observation and the target image, which were then fused into an Asynchronous Advantage Actor-Critic (A3C) \cite{A3C} model. Subsequent methods aimed to enhance navigation performance by incorporating techniques such as knowledge graphs \cite{Yang2019} or spatial relationships\cite{Druon2020}\cite{Ye2021b}\cite{Lyu2022}, large-scale training\cite{Wijmans2019}, learning from demonstrations \cite{Ramrakhya2022}, data augmentation \cite{Maksymets2021}, auxiliary tasks \cite{Ye2021a}, and zero-shot learning \cite{zson}\cite{cliponwheel}. These approaches seek to improve the efficiency of end-to-end navigation policies and enhance their generalization to new environments.
In contrast, modular methods, which have a hierarchical structure, demonstrate a stronger ability to capture long-term environmental features. These methods rely on a global policy to select waypoints on an environment map and use local path planners to execute navigation actions. The global policy is learned from various feature representations using reinforcement learning or supervised learning, such as topological graph \cite{Chaplot2020a}, geometry map \cite{Chaplot2020}, semantic map \cite{SemExp}, or potential functions \cite{poni}.

Recent advances in large pre-trained vision and language models have shown promise across domains, such as scene understanding\cite{layoutgpt} \cite{Conceptgraph}, semantic segmentation \cite{lseg}\cite{dino}, and visual navigation \cite{Yu2023a}\cite{VLMaps}. Researchers have explored using these models to guide robots in navigating unknown environments. For instance, CLIP on Wheels \cite{cliponwheel} adopts a straightforward approach where the robot explores the nearest frontier until the target object is detected using CLIP features. ZSON \cite{zson} leverages LLMs for zero-shot object navigation by extracting commonsense knowledge about relationships between targets and observed objects. Our previous work \cite{Yu2023a} proposed using LLMs to select the most promising frontiers, guiding the expansion of exploration areas and increasing the likelihood of finding the target object. Similarly, methods like ESC \cite{ESC} and LFG \cite{LFG} use multiple queries to LLMs to rank frontiers within the exploration map. Additionally, VLFM \cite{vlfm} employs BLIP-2 \cite{BLIP2} to compute the similarity between RGB images and prompt text, generating a language-grounded value map to select optimal exploration frontiers for zero-shot object category navigation.

Despite the success of reinforcement learning and vision-language model-based techniques in visual navigation tasks, they still struggle with processing language-descriptive targets that involve object attributes and spatial relationships. Moreover, most approaches suffer from limitations related to 2D exploration maps, which are inadequate for multi-floor environments. These methods lack a comprehensive 3D representation of the scene, limiting their applicability for tasks like planning, manipulation, or navigation.
In this work, we focus on exploring and locating specific objects using either category names or language descriptions in unknown environments. Our approach utilizes a 3D object-centric map and leverages the capabilities of large vision-language models to enhance the spatial reasoning performance of robots.

\subsection{Visual Target Navigation With Language Description}
Visual navigation involving not just object categories but also complex language-based descriptions presents a significant challenge for robots, as they must identify both the attributes and spatial relationships of the target within an unfamiliar environment. Prior works in language-based navigation often provide detailed step-by-step instructions for reaching a goal \cite{navgpt}\cite{discussnav}\cite{instructnav}, but they seldom utilize descriptive information about the objects themselves.
Recent advancements in large pre-trained vision-language models have shown their effectiveness in robotic navigation tasks, offering powerful tools for interpreting complex target descriptions. Some approaches use these models to create open-vocabulary semantic maps \cite{VLMaps}\cite{Conceptgraph}, ground language-embedded landmarks onto these maps, and then plan paths toward the target to complete the navigation task. For example, VLMaps \cite{VLMaps} employs the LSeg \cite{lseg} model to encode RGB images, aligning pixel embeddings with 3D map locations to form a semantic map. Similarly, OpenScene \cite{OpenScene} calculates dense features for 3D points, embedding them concurrently with text strings and image pixels in the CLIP feature space. In contrast, ConceptGraphs \cite{Conceptgraph} introduces an open-vocabulary 3D scene graph representation that assigns features exclusively to object nodes, significantly reducing storage requirements and enhancing scalability. Moreover, LLM-grounder \cite{LLMGrounder} utilizes large language models (LLMs) to break down complex natural language queries into semantic components and employs visual grounding tools such as OpenScene \cite{OpenScene} or LERF \cite{lerf2023} to identify targets in 3D scenes. LM-Nav \cite{lm-nav} takes a different approach by using GPT-3 to parse text instructions into landmarks, integrating predefined graphs with CLIP to ground these language-based landmarks. SayPlan \cite{2023sayplan} presents a method for large-scale task planning in robotics using 3D scene graph representations, leveraging the hierarchical structure of dynamic 3D scene graphs and implementing an iterative replanning pipeline with feedback from a scene graph simulator. Despite these innovative strategies, the reliance on pre-built semantic maps makes these methods unsuitable for robotic navigation in unknown environments. Additionally, the simultaneous use of multiple large-scale multimodal models and complex system designs hinders real-time performance, which is crucial for effective robotic navigation.

There are some related datasets that include language descriptions of target objects, such as \cite{scannet}\cite{sceneverse}, but these datasets primarily focus on grounding tasks. Goat-bench \cite{khanna2024goatbench} introduces the first multi-modal navigation dataset that incorporates language descriptions and provides a benchmark for universal navigation tasks. In this paper, we use the Goat-bench dataset \cite{khanna2024goatbench} to evaluate visual navigation performance with language-descriptive targets.

\subsection{Game-Theoretic Large Models}
Current vision-language models perform quite well in robotic tasks that involve generating or verifying factual assertions, owing to their strong capabilities in scene understanding and reasoning. However, these models are far from perfectly reliable, and recent research indicates that their tendency to generate false yet commonly repeated statements increases with model size \cite{hallucinate}.
For example, large-scale vision-language pre-trained models often hallucinate non-existent visual objects when generating text based on visual inputs. The likelihood of these hallucinations grows with the complexity and scale of the models, as the vast amounts of data they process can lead to the generation of common but incorrect outputs.
The situation is further complicated by the different affordances offered by vision-language models (VLMs) for handling factual generation tasks. VLMs can be used in generative modes (e.g., by predicting the most likely answer to a question) or discriminative modes (e.g., by assessing whether a given answer is acceptable for a posed question). These two approaches do not always yield consistent results: generative methods may struggle when the probability distribution is spread across multiple conflicting answers \cite{Wang2022}\cite{Mitchell2022}, while discriminative methods may fail due to miscalibration \cite{Han2023} or subtle variations in question-wording \cite{Jiang2020}. Consensus Game \cite{consensusgame} frames language model decoding as a regularized imperfect-information sequential signaling game, using equilibrium search to enhance the coherence of language models by balancing the generator and discriminator. 
Motivated by this feature, in this paper, we adopt a similar approach by designing both a generator and a discriminator but focus on applying game-theoretic vision-language models to the visual target navigation task. This approach aims to reduce hallucinations in VLMs and improve the accuracy of identifying visual targets.
To the best of our knowledge, this study represents the first attempt to integrate game-theoretic vision-language processing tools into robotic navigation tasks.


\begin{figure*}[htbp]
    \centering
    \includegraphics[scale=0.65]{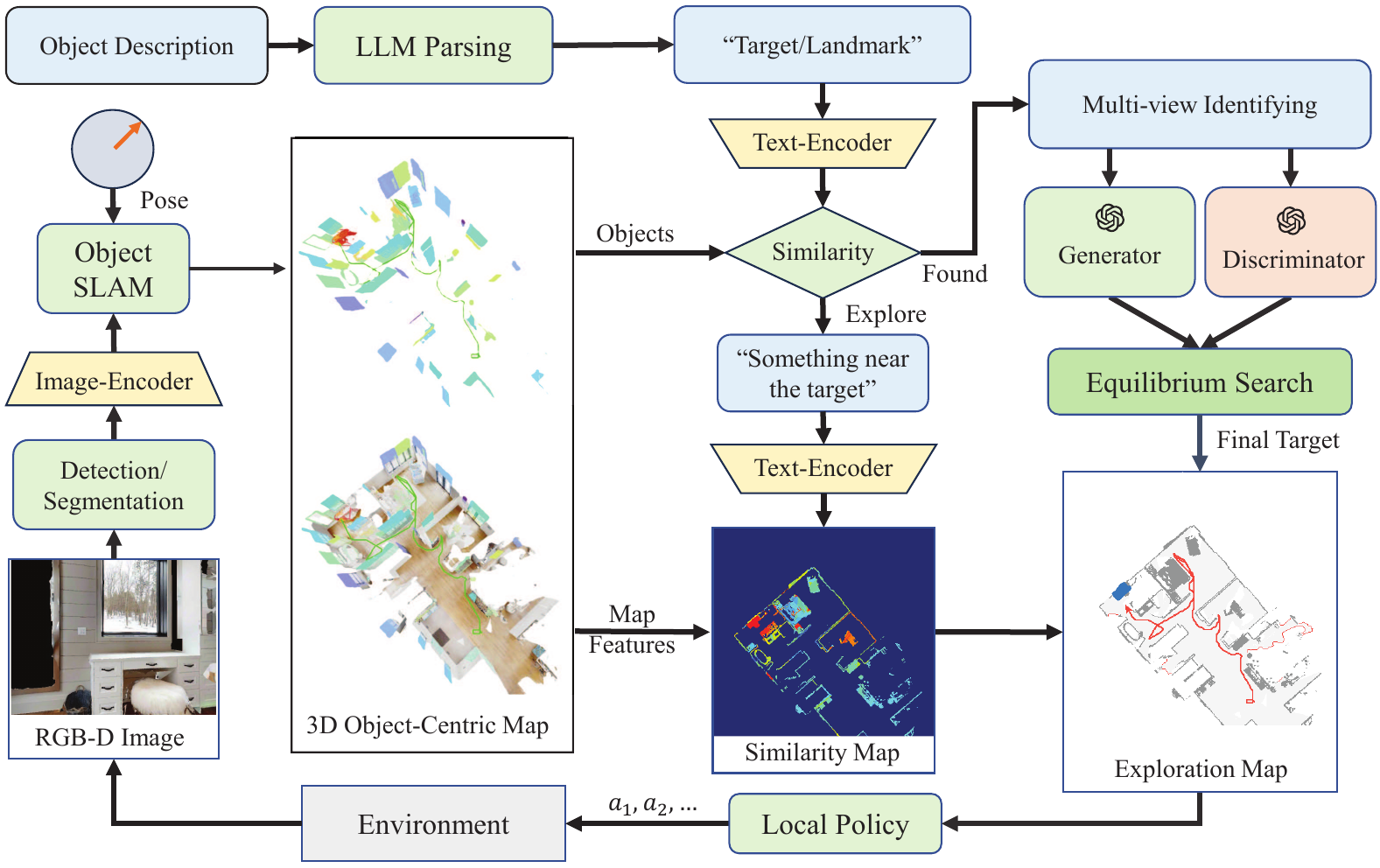}
    \caption{This framework utilizes posed RGB-D frames to generate a 3D object-centric map and an exploration map for robot navigation. Target descriptions are parsed by LLM to set the primary navigation goals. Using CLIP-based similarity assessments, the system evaluates the relevance between the target and environmental features to direct exploration activities. Upon detecting a potential target, a game-theoretic vision-language model analyzes spatial relationships described in the target instructions. Achievement of the long-term goal or target identification triggers a local policy that dictates the robot's final actions.}
    \label{fig:system_architecture}
    \vspace{-0.4cm}
\end{figure*}

\section{Method}

In this section, we define the visual target navigation task involving both object goals and language-based descriptions, and detail the navigation framework that incorporates a game-theoretic vision-language model.

\subsection{Task Definition}
\label{sec:task}
The visual target navigation task involves the agent navigating an environment to find an instance of the object category or a special language descriptive target. The query of the target with category or description set is represented by $Q = \left\{q_0, \dots, q_n\right\}$, and the scene can be represented by $S = \left\{s_0, \dots, s_n\right\}$.
Each episode begins with the agent being initialized at a random position $p_i$ in the scene $s_i$ and receives the target description $q_i$. Thus, an episode can be denoted as $E_i = \left\{s_i, q_i, p_i\right\}$.
At each time step, the agent gets the observation $o$ from the environment and takes an action $a \in A$. The observation $o$ includes RGB-D images $I$, the agent's location and orientation $p$, and the target object description $q$. The action space, denoted by $\mathcal{A}$, includes six actions: $move\_forward$, $turn\_left$, $turn\_right$, $look\_up$, $look\_down$, and $stop$. The $move\_forward$ action agent moves 25 cm, while the $turn\_left$, $turn\_right$, $look\_up$, or $look\_down$ actions rotate the agent 30 degrees. The $stop$ action is used when the agent believes it has found the target object. An episode is considered successful if the agent takes the $stop$ action when it is close enough to the target. Each episode has a maximum limit of 500 time steps.

\subsection{System Overview} 
Our framework is illustrated in Fig. \ref{fig:system_architecture}. 
Firstly, given a set of posed RGB-D frames, we use a class-agnostic segmentation model to identify candidate objects, associate them across multiple views using geometric and semantic similarity measures, and instantiate these objects in a 3D object-centric map. A 2D occupancy map is then generated by projecting the 3D map, which is used for exploration and planning. Frontiers are extracted from this occupancy map to define potential areas for further exploration.	
Secondly, the input object description is parsed by large language models (LLMs) to identify the primary goal without considering spatial relationships. A similarity map is then computed based on the CLIP similarities between all the objects in the environment and the target description. Using this similarity map, a long-term goal is selected from the frontiers to guide the agent in exploring and searching for the target.
If any related objects are detected based on the CLIP features, the game-theoretic vision-language model is employed to find the equilibrium between the Generator and Discriminator for all candidate objects. This process uses multi-view images to determine which candidate best matches the target description.
After identifying the long-term goal or locating the target object, the local policy plans a path and executes the final action to either continue exploring the environment or search for the target object.

\subsection{Map Construction}

\begin{figure*}[t]
    \centering
    \includegraphics[scale=0.70]{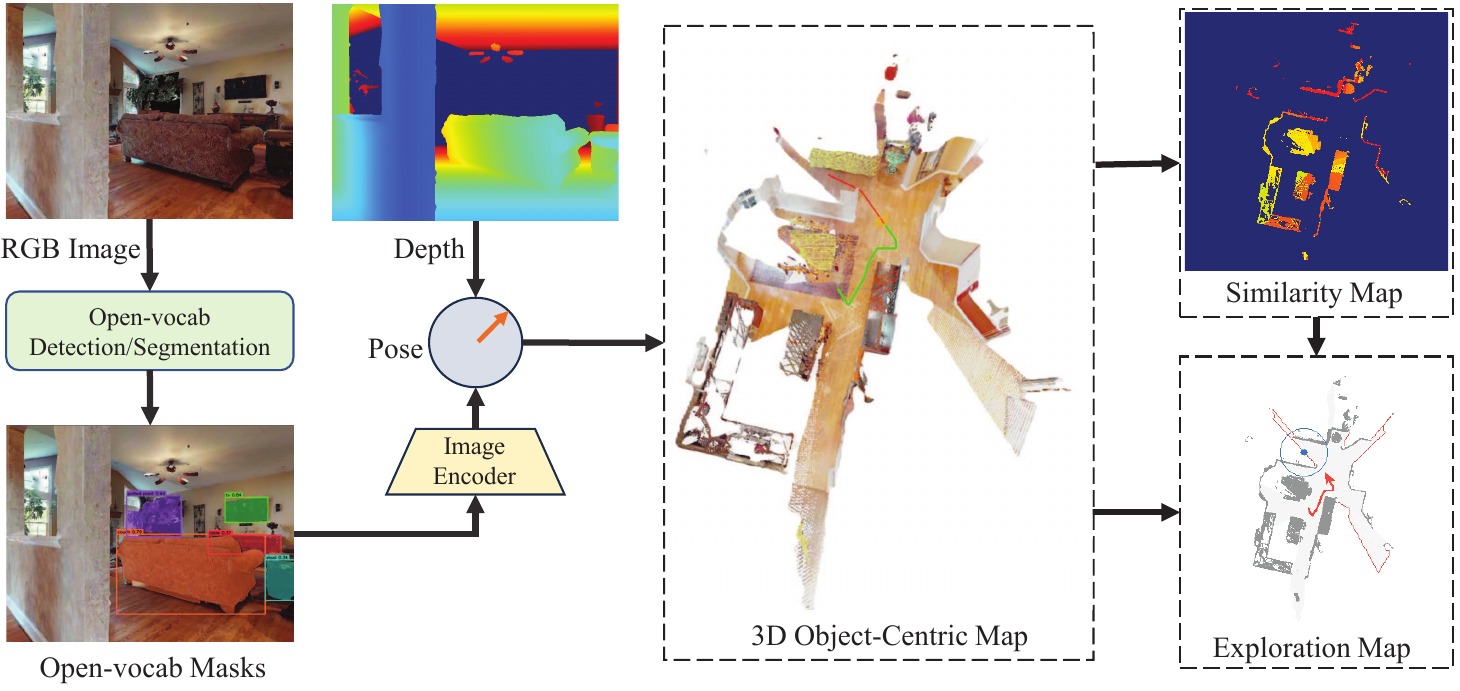}
    \caption{The process of the map building. The framework takes RGB-D images as input to generate a 3D object-centric map, in which each object in this map is represented as a group of point clouds and CLIP features. Based on the CLIP text encoder, the similarity map between our target and the objects in the map can be calculated. The exploration map is also obtained from the projection of the 3D point cloud map, which can be used to generate the frontiers and plan the paths.}
    \label{fig:mapping}
    \vspace{-0.4cm}
\end{figure*}

\subsubsection{3D Object-Centric Map}

Given a sequence of RGB-D images $\mathcal{I} = \{I_1, \dots, I_t\}$ and the pose of the agent $\mathcal{P} = \{p_1, \dots, p_t\}$, the 3D object-centric map $\mathcal{M}$ is built. 
The process of map generation can be seen in Fig. \ref{fig:mapping}.
Firstly, when processing a frame $I_t$, an open-vocabulary detection model $Det(\cdot)$ is used to detect bounding boxes for all candidate objects, and a class-agnostic segmentation model $Seg(\cdot)$ then extracts a set of masks corresponding to candidate objects in each bounding box. Each extracted mask is passed through a pre-trained CLIP visual encoder $Clip(\cdot)$ to obtain visual features. The depth information associated with each masked region is projected into 3D space using $F_{proj}(\cdot)$. This results in a point cloud and its corresponding unit-normalized semantic features for each object associated across multiple views with $F_{asso}(\cdot)$.
Secondly, for each newly detected object from subsequent frames, semantic and geometric similarities are calculated relative to all existing objects in the 3D map. The semantic similarity is evaluated using a variant of the CLIP features, while the geometric similarity is determined by the degree of overlap between the new object's geometry and existing objects. If the combined similarity exceeds a certain threshold, the new object is merged with the existing object; otherwise, it is added as a new object to the map.
This map $\mathcal{M}$ is built incrementally, incorporating each incoming frame $I$ and its associated pose $p$ by either merging with existing objects or instantiating new objects:
\begin{equation}
    \mathcal{M} = F_{asso} \{\sum_{\tau=1}^{t}  F_{proj}(Clip(Seg(Det(I_\tau))), p_\tau)\} 
\end{equation}

\subsubsection{2D Exploration Map}
To enable efficient exploration, we construct a 2D exploration map for frontier extraction and path planning. The 2D map is initialized with zeros at the start of each episode, with the agent placed at the center.
All relevant 3D point clouds are projected onto a top-down 2D map, which consists of the obstacle and explored channels. Frontiers are then extracted from these two channels. First, points above the floor in the 3D map are selected and projected onto the obstacle map, while all 3D points are projected to create the explored map. The explored edge is then identified by detecting the largest contours in the explored map.
Next, we generate the frontier map by dilating the edge of the obstacle map and calculating the difference between the explored map and the obstacle map. Finally, connected neighborhood analysis is used to identify and cluster frontier cells into chains. Clusters that are too small are discarded as frontiers.  The frontiers are shown in the exploration map of Fig. \ref{fig:mapping} with red slim lines. 

\subsection{Exploration Policy}
\label{sec:Policy}
\subsubsection{Frontier-Based Semantic Exploration}

After constructing the 3D object-centric and exploration maps, a frontier scoring policy is implemented to select target-relevant waypoints that can guide the agent in exploring the unknown environment more efficiently and locating the target.
The evaluation of frontiers is based on two types of scores: the geometry score and the semantic score.  
For the geometry score, we use the cost-utility approach proposed in \cite{Julia2012}. For each frontier cell $f \in F$, we can get the geometry score $S^{Geo}(f)$
\begin{equation}
    S^{Geo}(f)=U(f)-\lambda_{C U} C(f)
\end{equation}
where $U(f)$ is a utility function, $C(f)$ is a cost function and the constant $\lambda_{C U}$ adjusts the relative importance between both factors.

For the semantic score, we calculate the CLIP similarity between each object in the 3D object map and the target's CLIP text features. The similarity values are then projected onto a 2D semantic map, denoted as $\mathcal{M}_{obj\_sem}$. Additionally, we compute the CLIP similarity between the visual features of each frame and the target's language features to generate another 2D semantic map $\mathcal{M}_{img\_sem}$. To determine the semantic score for each frontier, we select a search window around the frontier and identify the highest CLIP similarity value within that window. For each frontier cell $f \in F$, we can get the semantic scores $S^{Sem\_o}(f)$ and $S^{Sem\_i}(f)$:
\begin{equation}
    S^{Sem\_o}(f) = Max \{\mathcal{M}_{obj\_sem}(f) \} 
\end{equation}
\begin{equation}
    S^{Sem\_i}(f) = Max \{\mathcal{M}_{img\_sem}(f) \}
\end{equation}
The final similarity map is the combination of the two similarities maps. We then select the frontier based on a score bound $B$ using the following strategy:
\begin{equation}
    f =\arg \max_{f_i} \left\{\begin{array}{cc}
        S^{Sem\_o}_{f_i},           & \text{if } \max\{S^{Sem\_o}\} > B_{sup}  \\[3mm]
        S^{Sem\_i}_{f_i}, & \text{elif } \max\{S^{Sem\_i}\}  > B_{inf}  \\[3mm]
        S^{Geo}_{f_i},            & \text{else}  
    \end{array}\right.
\end{equation}
where the $B_{sup}$ and $B_{inf}$ are the supremum and infimum of bound $B$.  
The merged similarity map can be seen in Fig. \ref{fig:mapping}.

\subsubsection{Waypoint Navigation} 

After initializing the map, the robot is provided with either a frontier waypoint or a target object waypoint to navigate toward, depending on whether a target object has been detected. To plan the path from the agent's current location to the waypoint, we use the Fast Marching Method (FMM) \cite{fmm}. The agent then selects a local goal within a limited range of its current position and executes the final action $a \in A$ to reach it. At each step, the local map and goal are updated incrementally with new observations. 

Navigating multi-floor scenes in the simulation presents a significant challenge for locating a specific object. In some cases, the target object and the agent's initial position may be on different floors. To find the object, the agent may need to traverse stairs, which affects the process of building the 2D exploration map. To deal with this problem, we design a strategy using the advantage of the 3D map that when no more frontier can be detected in the 2D exploration map, which means that the current floor has been explored, the robot needs to decide to go upstairs or downstairs for further exploration based on the current environment and the attributes of the target. For instance, if the robot cannot find a bed on the current floor, it should go upstairs, as bedrooms are typically located on higher floors. Conversely, for items such as a TV, sofa, or oven, descending to a lower floor is often a better option. In our framework, a large language model is used to determine the appropriate direction for further exploration when navigating stairs.

The current exploration policy is designed specifically for object goal search in visual target navigation tasks. In the next section, we will introduce how this policy can be adapted for language descriptive targets.

\subsection{Visual Target Identifying}
\subsubsection{Object Goal Identifying}
For visual target navigation based on an object category's name, identification can be achieved using the similarity values of objects stored in the explored semantic map $M_{obj\_sem}$. Leveraging the powerful capabilities of the CLIP model, we can also handle open-vocabulary object descriptions with specific attributes, such as ``black single sofa." However, CLIP struggles to maintain consistent similarity values from different views of the same object, particularly when the observed view is not close enough. To address this challenge, we implemented a candidate list for potential targets. When an object has a reasonably high similarity value in $M_{obj\_sem}$ but does not exceed a specified threshold, it is added to the candidate list and marked as a temporarily found target. The robot then moves closer to the candidate object to obtain a better view, allowing the similarity value to be updated in $M_{obj\_sem}$ with greater confidence and compared against the threshold.     
However, for language-descriptive targets that involve spatial relationships, CLIP struggles to handle these cases effectively \cite{Conceptgraph}\cite{LLMGrounder}, as it primarily models orderless content and does not adequately account for relations and the order of elements when processing text and visual information.

\subsubsection{Language Goal Identifying}

\begin{figure}[t]
    \centering
    \includegraphics[scale=0.4]{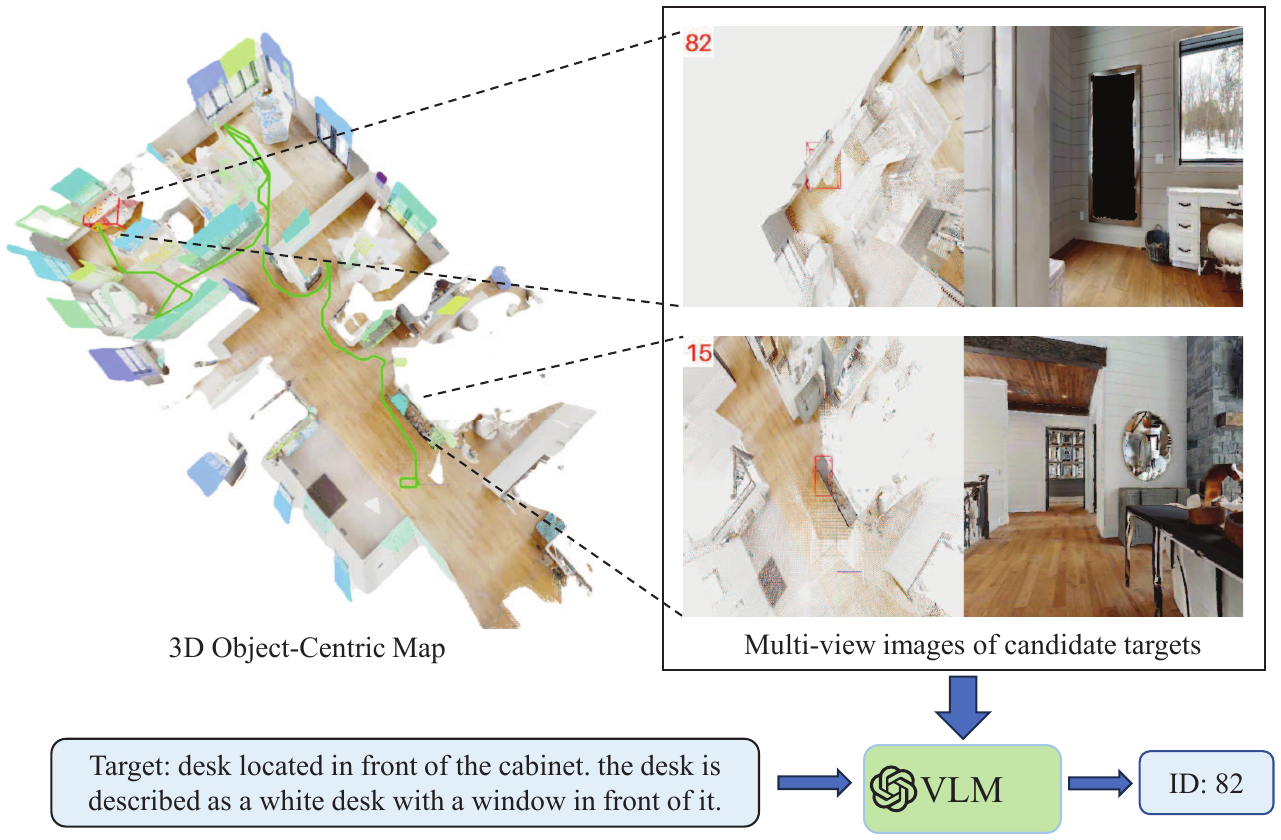}
    \caption{A case of identifying a white desk in the scene. There are two candidate targets detected during the navigation. Based on the multi-view images and the language description of the target, the vision language model can output the target's ID from the candidate list that most matches the description.}
    \label{fig:mvi}
\end{figure}

To address the challenge of identifying language-descriptive targets during exploration, we propose a two-step approach:

First, a target with attributes extracted from the language description by large language models. When the language description is provided, the LLM extracts the main goal and its attributes, while ignoring spatial relationships, which are handled separately by the CLIP-based 3D object-centric map. For example, in the description ``the black office chair between the screen and table", the main goal is the ``black office chair". If the description does not include spatial relationships, the extracted goal is used directly for navigation. Once the main goal is determined, the exploration policy outlined in Section \ref{sec:Policy} is employed to search for the target.

Second, spatial relationships specified in the description are verified using vision-language models once the main object has been identified based on the CLIP-based 3D map. To understand spatial relations in the 3D environment, we propose a multi-view approach using the 3D object-centric map $\mathcal{M}$. The target's top-view image $I_{top}$ and first-person view image $I_{first}$ are extracted from $\mathcal{M}$. The first-person view image helps verify the attributes and local spatial relationships of the target object, while the top-view image compensates for situations where the agent is too close to the object, making it difficult to assess the surrounding layout.
Each time the robot detects a new candidate target, the top-view and first-person view images of all candidate objects are provided to the vision-language model along with the language description. The model analyzes the combined input images $I_m = I_{top} + I_{first}$ and identifies the candidate that best matches the language description. If a match is found, the model outputs the object ID. If no candidates match the description, all current candidates are rejected, and the robot continues exploring the environment until a new target is detected. This approach enables the robot to explore the environment effectively and identify complex targets, especially in cases where multiple instances of the main goal exist.

Although vision-language models assist in identifying language descriptive targets, even minor errors or hallucinations in decision-making can negatively impact the success of navigation, such as wrongly identifying the target or misunderstanding the spatial relationships. In this paper, we address this issue by employing game-theoretic vision-language models to enhance the robustness of the identification process.

\subsection{Vision-Language Equilibrium Search}

\begin{algorithm}[!b]
    \caption{Vision Language Equilibrium Search}\label{alg:vl_equilibrium_search}
    \textbf{Input:} 
    \begin{itemize}
      \item $\mathcal{I}_m= \left\{I^0_m, \dots, I^N_m\right\}$: the candidate target set of multi-view images;
      \item $q$: target language description;
      \item $r \in(0, N)$: candidate target ID;
      \item $\eta_G, \eta_D > 0$: learning rate hyperparameter;
      \item $iter$: number of iterations;
      \item $bi$: bias for initial policies.
    \end{itemize}
    \textbf{Output:} 
    \begin{itemize}
      \item $\pi_G^*$: Generator;
      \item $\pi_D^*$: Discriminator.
    \end{itemize}
    
    \textbf{Algorithm:} 
    \begin{algorithmic}[H]
        \Function{Initialize}{}
          \State $ \pi^{(1)}_G \gets {P_{VLM}(r \mid \mathcal{I}_m, q)} + bi$ 
          \State $ \pi^{(1)}_{\mathcal{D}} \gets P_{VLM}(I^r_m, q) / \sum_{r'} P_{VLM}(I^{r'}_m, q) + bi $
        \EndFunction
        
        \Function{Play}{}
          \State \textbf{update each policy}
          \For{each iteration $t \in range(iter)$}
            \State $Q^{(t)}_G += \frac{1}{2t} \sum_{\tau=1}^{t} \pi^{(\tau)}_D $
            \State $ Q^{(t)}_D += \frac{1}{2t} \sum_{\tau=1}^{t} \pi^{(\tau)}_G $
            \State $\pi^{(t+1)}_G \propto \exp \left\{ \frac{Q^{(t)}_G + \lambda_G \log \pi^{(1)}_G} {\left( 1 / {\eta_G t} + \lambda_G \right)} \right\}$
            \State $\pi^{(t+1)}_D \propto \exp \left\{ \frac{Q^{(t)}_D + \lambda_D \log \pi^{(1)}_D}  {\left( 1 / {\eta_D t} + \lambda_D \right)} \right\}$
            
          \EndFor
        \EndFunction
        
    \end{algorithmic}
\end{algorithm}

To address the hallucination challenge faced by vision-language models in this task, we present an approach that reconciles generative and discriminative VLM decoding procedures by formulating decoding as a signaling game. Unlike the method proposed in \cite{consensusgame}, our approach emphasizes searching for the equilibrium of the game to enhance the coherence of vision-language models, thereby improving accuracy in navigation tasks.

\subsubsection{Vision-Language Model as Equilibrium Search} 

We study the problem of obtaining correct output from a vision-language model, which maps input the observation $x$ to output result $y$ according to special distribution $P_{VLM} (y | x)$. 
While the techniques presented here are general, this paper focuses on identifying language-descriptive targets based on a query $q$ and a set of candidate targets $ \mathcal{I}_m = \left\{I^0_m, \dots, I^N_m\right\}$ which may themselves have been identified from the 3D object-centric map. Given a set of candidates, the target id $r \in(0, N)$ can be selected using VLMs in two ways:
\begin{itemize}
    \item \textbf{Generatively ($\pi_G$): } By providing the descriptive query $q$, the set of candidates $ \mathcal{I}_m = \left\{I^0_m, \dots, I^N_m\right\}$, and a natural language prompt indicating that a matched answer is desired. In this approach, the VLM models a distribution $P_{VLM} (r|\mathcal{I}_m, q)$.
    \item \textbf{Discriminatively ($\pi_D$): } By providing the descriptive query $q$, a possible candidate image $I^r_m$, and a prompt asking for the correctness of the candidate. Here, the VLM models a distribution $P_{VLM} (I^r_m, q)$.
\end{itemize}
At the start of the game, the target description $q$ and all the candidate images $ \mathcal{I}_m$ are observed by the Generator $\pi_G$ which generates an answer from the fixed set of candidates IDs $r \in(0, N)$ that perfectly matches the description. Discriminator $\pi_D$ then receives the description $q$ and the generated answer $I^r_m$, and attempts to determine the correctness of the generated answer. Both players receive a payoff of 1 if the $\pi_{\mathcal{D}}$ confirms that the generated answer is correct, and 0 otherwise.

With this definition, the players' expected utilities can be determined using the Nash equilibrium of the game:

\begin{equation}
    u_G(\pi_G, \pi_D) :=u_{P}(\pi_G) + \frac{1}{2} \sum_{r \in(0, N)} \pi_G(r|\mathcal{I}_m, q) \cdot \pi_D(I^r_m, q), 
\end{equation}
\begin{equation}
    u_D(\pi_G, \pi_D) := u_{P}(\pi_G) + \frac{1}{2} \sum_{r \in(0, N)} \pi_G(r|\mathcal{I}_m, q) \cdot \pi_D(I^r_m, q),
\end{equation}
where $u_{P}(\pi_G)$ and $u_{P}(\pi_G)$ are penalty terms applied to the Generator $\pi_G$ and the Discriminator $\pi_D$, respectively, to discourage them from deviating significantly from their initial policy pair. This approach is inspired by \cite{consensusgame}:
\begin{align}
    u_{P}(\pi_G) & := -\lambda_G \cdot D_{KL}[\pi_G(\cdot \mid \mathcal{I}_m, q), \pi^{(1)}_G(\cdot \mid \mathcal{I}_m, q)] , \\
    u_{P}(\pi_D) & := -\lambda_D \cdot D_{KL}[\pi_D(I^r_m, q), \pi^{(1)}_D(I^r_m, q)],
\end{align}

\subsubsection{Equilibrium Search}
In this section, we present Algorithm \ref{alg:vl_equilibrium_search}, which provides a detailed procedure for performing no-regret learning in this game to obtain the consensus policies. Notably, this approach focuses on modifying only the signaling policies $u_G$ and $u_D$.

\textbf{Initial policies: } At time \( t = 1 \), the initial policies are defined as \(\pi^{(1)}_G\) and \(\pi^{(1)}_D\) of the Generator and Discriminator, respectively, as follows. \(\pi^{(1)}_G\) normalizes \(P_{VLM}\) across \(r\):
\begin{equation}
    \pi^{(1)}_{\mathcal{G}}(r \mid I_m, q) \propto P_{VLM}(r \mid \mathcal{I}_m, q).
\end{equation}
Similarly for the Discriminator, the initial policy normalizes across \(r\):
\begin{equation}
    \pi^{(1)}_{\mathcal{D}}(I^r_m, q) \propto \frac{P_{VLM}(I^r_m, q)}{\sum_{r'} P_{VLM}(I^{r'}_m, q)}.
\end{equation}
This step is used to extract a well-calibrated Generator and Discriminator from $P_{VLM}$. We can get the distribution using the related prompts from VLM and sample them separately.

\begin{figure}[htbp]
    \centering
    \includegraphics[scale=0.43]{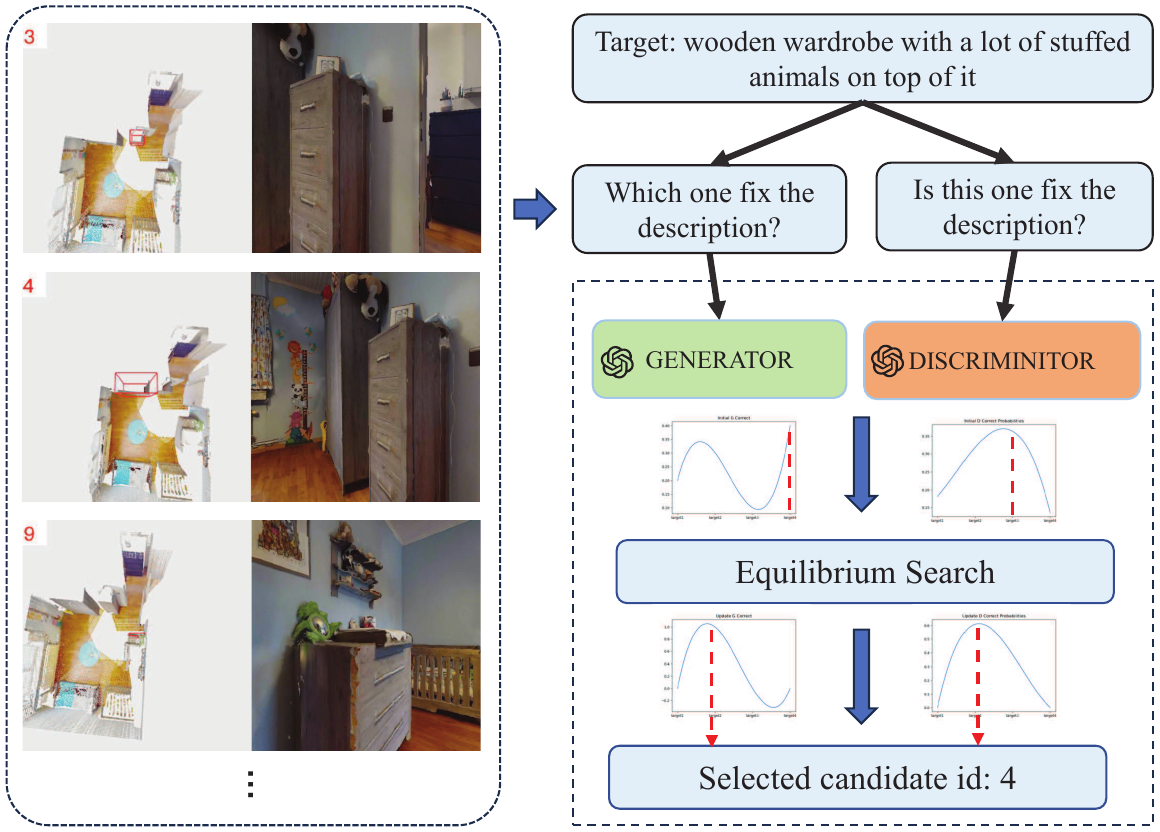}
    \caption{A case of Nash equilibrium search based on vision language models. After getting all candidate targets, the Generator and Discriminator can be used to infer a coherent result.}
    \label{fig:game}
\end{figure}

\textbf{Update policies: }
A well-known result in the theory of imperfect-information sequential games is that minimizing regret can be achieved by solving separate, local regret minimization problems at each information set of the game. In this task, the local regret minimization problems consist of a bilinear term combined with a strongly convex KL-regularization term. These composite utilities can be addressed using the piKL algorithm \cite{Jiang2020}, which is specifically designed for regret minimization in KL-regularized settings. In our approach, piKL requires each player to keep track of their average values:
\begin{equation}
    Q^{(t)}_G(r \mid \mathcal{I}_m, q) := \frac{1}{2t} \sum_{\tau=1}^{t} \pi^{(\tau)}_D(I^r_m,q),
\end{equation}
\begin{equation}
    Q^{(t)}_D(I^r_m,q) := \frac{1}{2t} \sum_{\tau=1}^{t} \pi^{(\tau)}_G(r \mid \mathcal{I}_m, q).
\end{equation}
Each player then updates their policy according to:
\begin{align}
    &\pi^{(t+1)}_G(r \mid \mathcal{I}_m, q)  \propto \notag \\ 
    &\exp \left\{ \frac{Q^{(t)}_G(r \mid \mathcal{I}_m, q) + \lambda_G \log \pi^{(1)}_G(r \mid \mathcal{I}_m, q)} {\left( 1 / {\eta_G t} + \lambda_G \right)} \right\},\\
    &\pi^{(t+1)}_D(I^r_m,q) \propto \notag \\
    &\exp \left\{ \frac{Q^{(t)}_D(I^r_m,q) + \lambda_D \log \pi^{(1)}_D(I^r_m,q)}  {\left( 1 / {\eta_D t} + \lambda_D \right)} \right\},
\end{align}
where $\eta_G, \eta_D > 0$ are learning rate hyperparameters, $\lambda_G, \lambda_D$ are used to guarantee the updated policies are within a proportional radius centered in the initial policies.

piKL no-regret dynamics \cite{Jiang2020} are known to have strong guarantees of convergence to an equilibrium point.
At convergence, vision language equilibrium search returns $\pi_G^*$ and $\pi_D^*$, which are the updated Generator and Discriminator policies. The final target object is then selected based on these updated policies. Fig. \ref{fig:game} illustrates an example of identifying the most relevant candidate target using the game-theoretic vision-language model. After obtaining all the multi-view images of the candidate targets, the Generator and Discriminator distributions are initialized with different prompts. The updated consensus policies are calculated using the equilibrium search strategy, resulting in distributions that are more coherent than the initial policies. Based on the $\pi_G^*$ and $\pi_D^*$, the final target is selected.

\section{Experimental Results}
\label{sec:result}

In this section, we evaluate the performance of our visual target navigation framework in a simulation environment with object goal and language description datasets. Additionally, we apply our method in a quadruped robot platform to validate its practicality for navigation tasks in real-world complex environments.

\subsection{Simulation Experiment}

\begin{figure*}[htbp]
    \centering
    \subfloat[step 10]
    {
        \includegraphics[width=4.25cm]{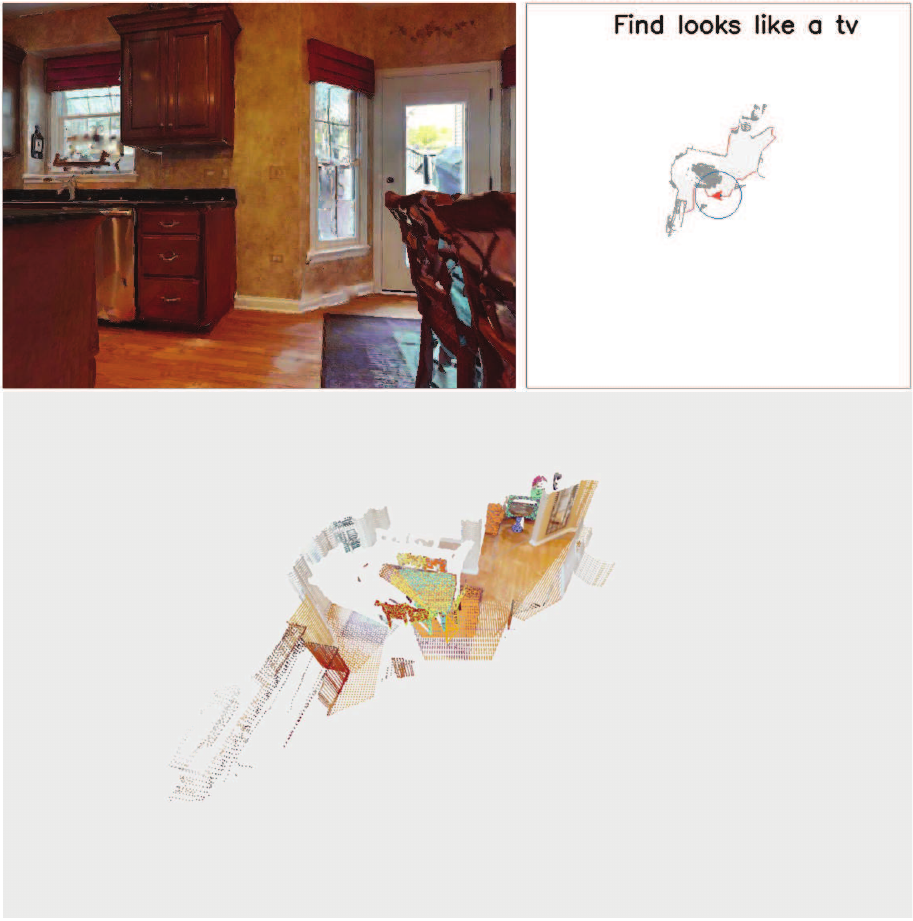}
    }
    \subfloat[step 40]
    {
        \includegraphics[width=4.25cm]{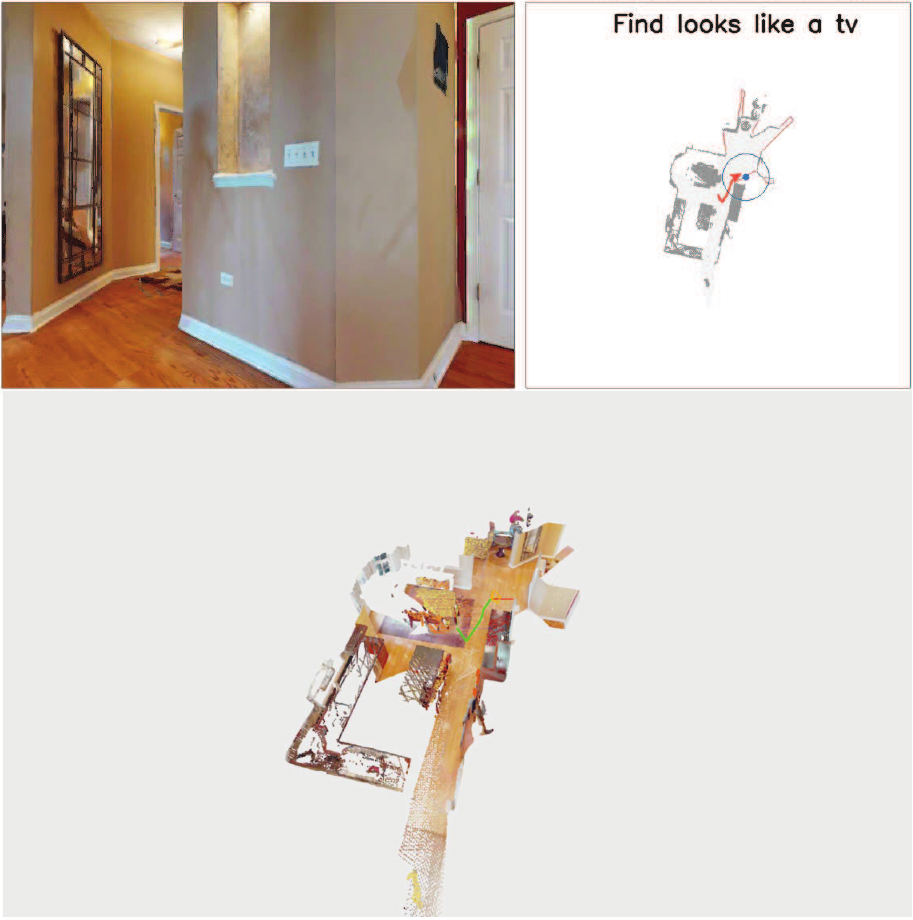}
    }
    \subfloat[step 70]
    {
        \includegraphics[width=4.25cm]{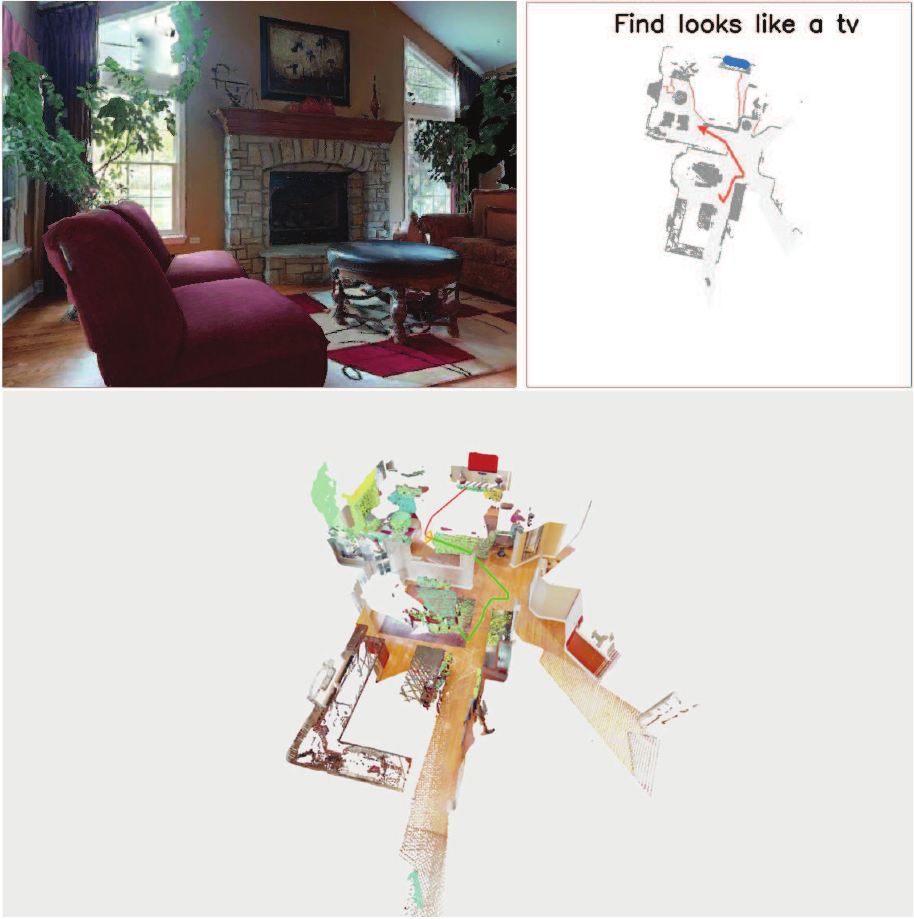}
    }
    \subfloat[step 100]
    {
        \includegraphics[width=4.25cm]{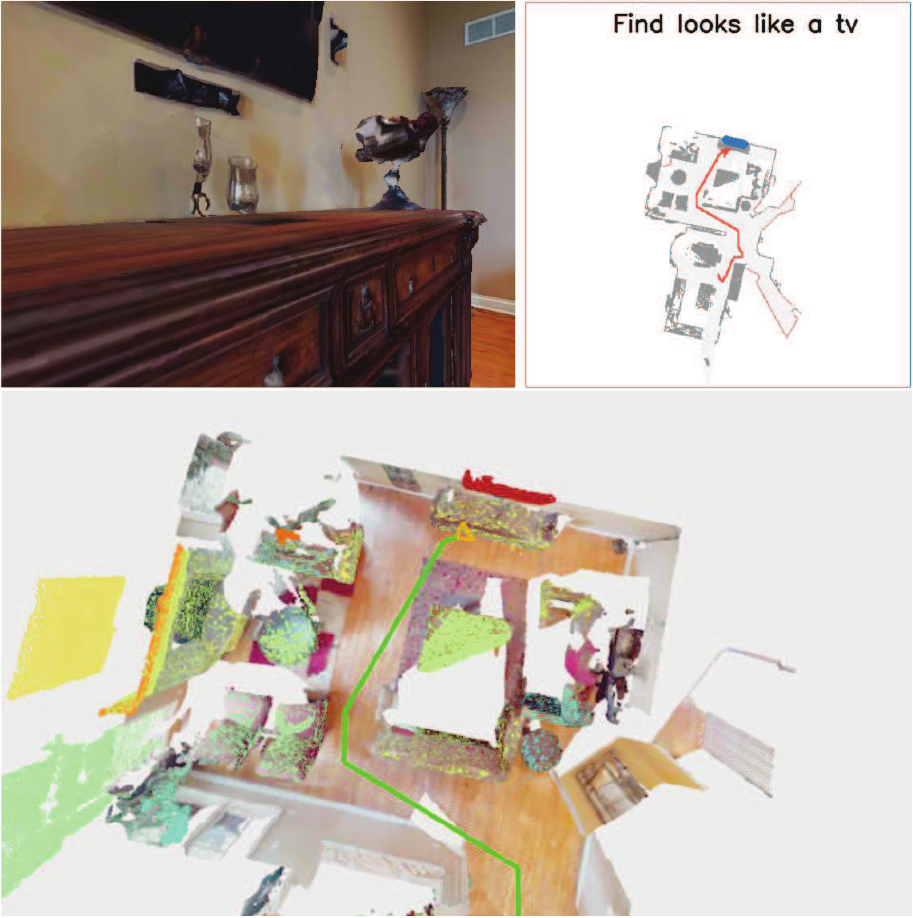}
    }
    \caption{The visual target navigation experiment process in the Habitat platform for finding a TV. There are the first-person view RGB image, exploration map, and 3D object-centric map. In the exploration map, the gray channel represents the barrier, the blue spot and the circle denote the long-term goal selected by our policy, the red thick line represents the trajectory of the robot, and the red thin line denotes the frontiers. In the 3D object-centric map, the green line represents the robot's trajectory, and the red line shows the planned path.}
    \label{fig:sim-ogn}
\end{figure*}

\subsubsection{Dataset}
We use high-resolution photorealistic 3D reconstructions from the HM3D dataset \cite{HM3D} to evaluate our framework. For navigation tasks based on category names, we use 20 validation scenes from the HM3D dataset, comprising 2,000 episodes across these scenes. There are six object goal categories defined in \cite{SemExp}: chair, couch, potted plant, bed, toilet, and TV.
For language-descriptive targets, we extract 316 episodes from Goat-bench \cite{khanna2024goatbench} to create the evaluation dataset. This dataset includes open-vocabulary language targets, incorporating attributes and spatial relationships. Each episode consists of a scene ID, one target (either a category name or language description), and a randomly selected starting position as defined in Section \ref{sec:task}.

\subsubsection{Implementation Details}
We conducted our evaluation on the 3D indoor simulator Habitat platform \cite{aihabitat}, using an observation space consisting of 480 × 640 RGB-D images, a base odometry sensor, a goal object represented as an integer, and a language description as a query. We use the DINO\cite{dino} as the open-vocabulary detection model $Det(\cdot)$ and Mobile-SAM\cite{SAM} as the class-agnostic segmentation model $Seg(\cdot)$.
Our implementation was based on publicly available code from Conceptgraph\cite{Conceptgraph} and L3MVN\cite{Yu2023a}, using the PyTorch framework. We set the score bound for the frontier as $B=[0.22, 0.26]$, and use the GPT4o-2024-05-13\cite{openai2023gpt4} and related versions of GPT4o-mini as the vision-language models. To initialize the policies $\pi^{(1)}_G$ and $\pi^{(1)}_D$, we sampled five times from the vision-language model and normalized the results.  We set $\eta_{\mathcal{G}} = \eta_{\mathcal{D}} = \lambda_{\mathcal{G}} = \lambda_{\mathcal{D}} = 0.1$ for the equilibrium search strategy, and configured the temperature parameter of the OpenAI API to 1.0 to facilitate thorough exploration of the initial policies.
We also utilized Open3D for visualizing the 3D object-centric map and the robot's state. An interactive interface was designed to visualize the map, display the robot's observations, and dynamically set the target description based on the constructed map.

\subsubsection{Evaluation Metrics}
We follow the evaluation metrics from \cite{Anderson2018}, using Success Rate (SR), Success weighted by Path Length (SPL), and Distance to Goal (DTG). SR is defined as $\frac{1}{N} \sum_{i=1}^{N} S_{i}$, and SPL is defined as $\frac{1}{N} \sum_{i=1}^{N} S_{i} \frac{l_{i}}{\max \left(l_{i}, p_{i}\right)}$, where $N$ is the number of episodes, $S_{i}$ is 1 if the episode is successful, else is 0, $l_{i}$ is the shortest trajectory length between the start position and one of the success position, $p_{i}$ is the trajectory length of the current episode $i$. The DTG measures the distance between the agent and the target goal at the end of the episode.

\subsubsection{Object Goal Navigation on HM3D}

\renewcommand\arraystretch{1.4}
\begin{table}[ht]
    \centering
    \fontsize{9}{8}\selectfont
    \begin{threeparttable}
        \caption{Results of Comparative Study in Object Goal Navigation.}
        \label{tab:obj_performance}
        \setlength{\tabcolsep}{2mm}{}
        {
            \begin{tabular}{ccccc}
                \toprule
                   {\bf Method}                      &{\bf Zero-Shot}   &{\bf Target}    &{\bf Success}      & {\bf SPL}        \cr
                \midrule
                SemExp \cite{SemExp}              & & ObjNav & 0.379       & 0.188                  \cr
                CoW \cite{cliponwheel}                                 & & ObjNav & 0.320       & 0.181                  \cr
                ZSON \cite{zson}                               &\checkmark  &ObjNav &  0.255 &  0.126       \cr
                ESC \cite{ESC}                                &\checkmark & ObjNav &  0.392 & 0.223        \cr
                L3MVN \cite{Yu2023a}                              &\checkmark & ObjNav &  0.504 & 0.231        \cr
                VLFM \cite{vlfm}                               &\checkmark  &ObjNav  & 0.525 &  0.304       \cr
                InstructNav \cite{instructnav}                        & \checkmark & Generic        & 0.580 &  0.209       \cr
                Ours                     &\checkmark  &Generic  & {\bf 0.613} & {\bf 0.268}        \cr
                \bottomrule
            \end{tabular}
        }
        
    \end{threeparttable}
\end{table}

\renewcommand\arraystretch{1.4}
\begin{table}[htbp]
    \centering
    \fontsize{9}{8}\selectfont
    \begin{threeparttable}
        \caption{Results of Ablation Study in Object Goal Navigation.}
        \label{tab:ablation_study}
        \setlength{\tabcolsep}{3mm}{}
        {
            \begin{tabular}{cccccc}
                \toprule
                \multicolumn{3}{c}{Ablation} & \multicolumn{2}{c}{ HM3D results }\cr
                \cmidrule(lr){1-3} \cmidrule(lr){4-5} 
                Near             & Img-sim       & Obj-sim       & Success $\uparrow$   & SPL $\uparrow $   \cr
                \midrule
                $\checkmark$        &               &              & 0.604                 & 0.243             \cr
                                    & $\checkmark$  &              & 0.607                 & 0.252             \cr
                                    &               & $\checkmark$ & 0.615                 & 0.264             \cr
                                    & $\checkmark$  & $\checkmark$ & 0.613                 & 0.268             \cr
                \bottomrule
            \end{tabular}
        }
    \end{threeparttable}
\end{table}

We compare our method with state-of-the-art object goal navigation techniques on HM3D datasets: SemExp \cite{SemExp}, CoW\cite{cliponwheel}, ZSON\cite{zson}, ESC\cite{ESC}, L3MVN\cite{Yu2023a}, VLFM\cite{vlfm}, and InstructNav\cite{instructnav}. SemExp\cite{SemExp}  is the first modular approach to address the object navigation task, using an explicit spatial map as scene memory and learning semantic priors through deep reinforcement learning to determine the next action. CoW\cite{cliponwheel} explores the nearest frontier until the target object is detected, utilizing either CLIP features or an open-vocabulary object detector. ZSON\cite{zson} is an open vocabulary method that uses CLIP to transfer a method for a different navigation task (ImageNav) zero-shot to the ObjectNav task. ESC\cite{ESC}, L3MVN\cite{Yu2023a}, and VLFM\cite{vlfm}  all employ semantic frontier-based exploration techniques, leveraging pre-trained language models or vision-language models to identify the most relevant frontier based on the features of the target and the map. InstructNav\cite{instructnav}, on the other hand, uses large language models (LLMs) to formulate a navigation plan based on instructions and combines multi-sourced value maps to select the optimal exploration waypoint.

Table \ref{tab:obj_performance} shows that our framework outperforms all learning-based and LLM-based zero-shot ObjectNav methods in terms of success rate, with a significantly higher SPL compared to InstructNav, which also achieves a high success rate. This demonstrates the considerable advantage of our framework in visual target navigation tasks using category names. A case process of finding a TV is illustrated in Fig \ref{fig:sim-ogn}.

To assess the relative importance of the various modules within our framework, we perform the ablations using the HM3D dataset: the nearest frontier module (Near), the image-based similarity map $\mathcal{M}_{img\_sem}$, and the object-based similarity map $\mathcal{M}_{obj\_sem}$, separately. The ablation study in TABLE \ref{tab:ablation_study} shows that our complete model achieves the best efficiency (row 4, TABLE \ref{tab:ablation_study}). Removing the Img-sim can also achieve great performance in success rate, but decrease the SPL (row 3, TABLE \ref{tab:ablation_study}). Removing the Obj-sim (row 2, TABLE \ref{tab:ablation_study}) decreases both the success rate and SPL, and replacing the similarity map with the nearest frontier policy (row 1, TABLE \ref{tab:ablation_study}) results in a further drop in performance, highlighting the crucial role of semantic similarity maps in our framework.

\subsubsection{Language Goal Navigation on HM3D}

\renewcommand\arraystretch{1.4}
\begin{table*}[htbp]
    \centering
    \fontsize{9}{8}\selectfont
    \begin{threeparttable}
        \caption{Results of Comparative Study in Language Goal Navigation.}
        \label{tab:vln_performance}
        \setlength{\tabcolsep}{6mm}{}
        {
            \begin{tabular}{ccccc}
                \toprule
                {\bf Method}		&{\bf VLM}        & {\bf Success}   & {\bf SPL}        & {\bf DTG}      \cr
                \midrule
                CLIP-Based \cite{clip}         &CLIP               & 0.253           & 0.106            & 4.697        \cr
                VLN-Generator \cite{openai2023gpt4}   	&GPT4o-mini         & 0.304           & 0.108            & 4.474         \cr	
                VLN-Ranking \cite{ranking}   	&GPT4o-mini         & \textbf{0.332}           & \textbf{0.113}            &\textbf{ 4.262}           \cr
                VLN-Game            &GPT4o-mini         & 0.354           & 0.118            & 4.119             \cr
                VLN-Generator\cite{openai2023gpt4}  	&GPT4o           	& 0.351           & 0.120            & 4.220            \cr    			
                VLN-Ranking \cite{ranking}   	&GPT4o          	& 0.364           & 0.124            & 4.335           \cr
                VLN-Game            &GPT4o         		& \textbf{0.367}           & \textbf{0.132}            & \textbf{4.093}             \cr
                \bottomrule
            \end{tabular}
        }
        
    \end{threeparttable}
\end{table*}

\begin{figure*}[htbp]
\centering
\subfloat[step 10]
{
    \includegraphics[width=4.25cm]{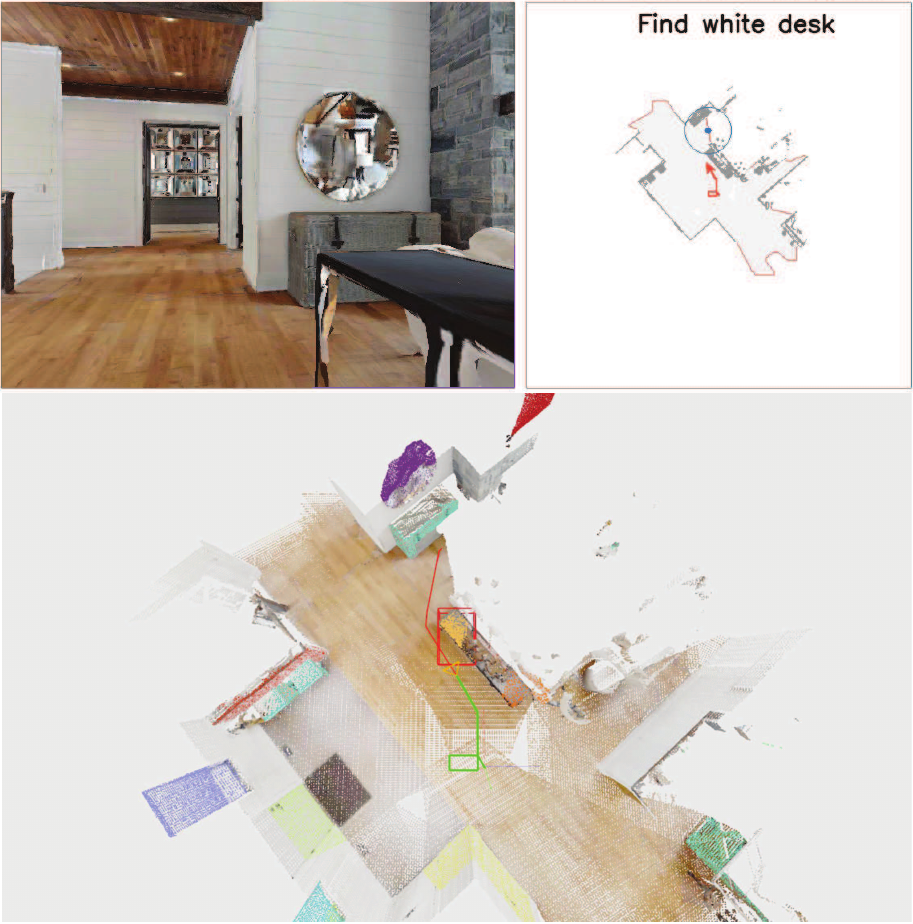}
}
\subfloat[step 40]
{
    \includegraphics[width=4.25cm]{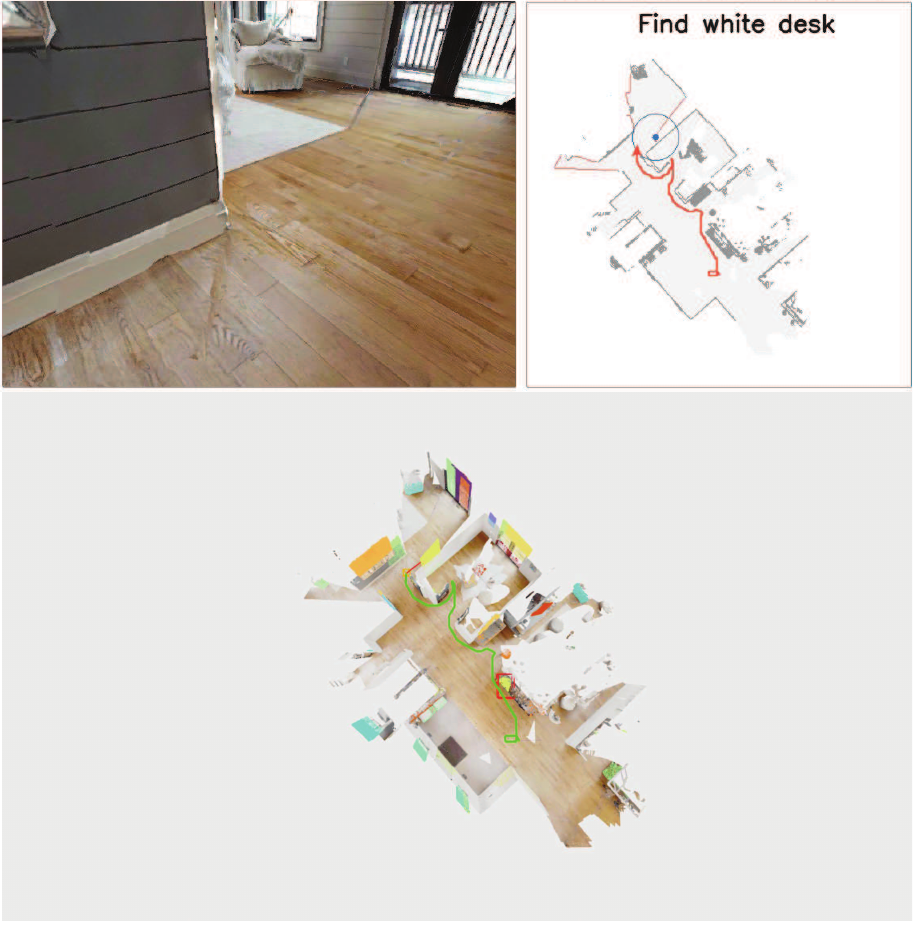}
}
\subfloat[step 70]
{
    \includegraphics[width=4.25cm]{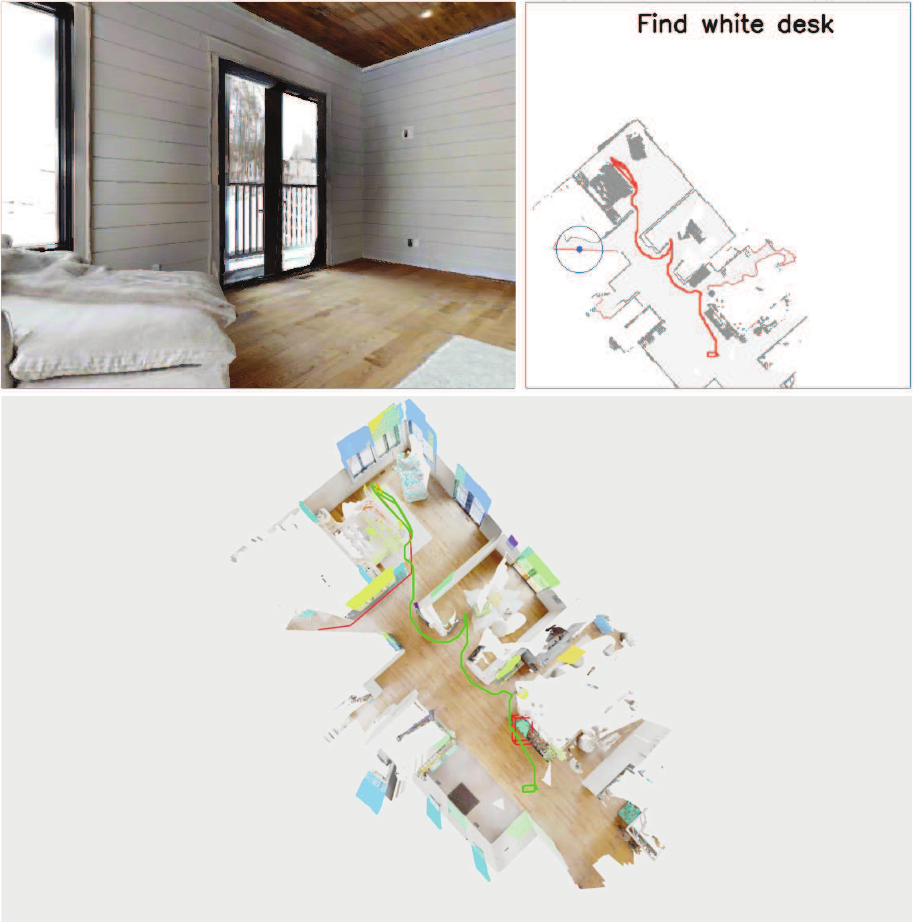}
}
\subfloat[step 100]
{
    \includegraphics[width=4.25cm]{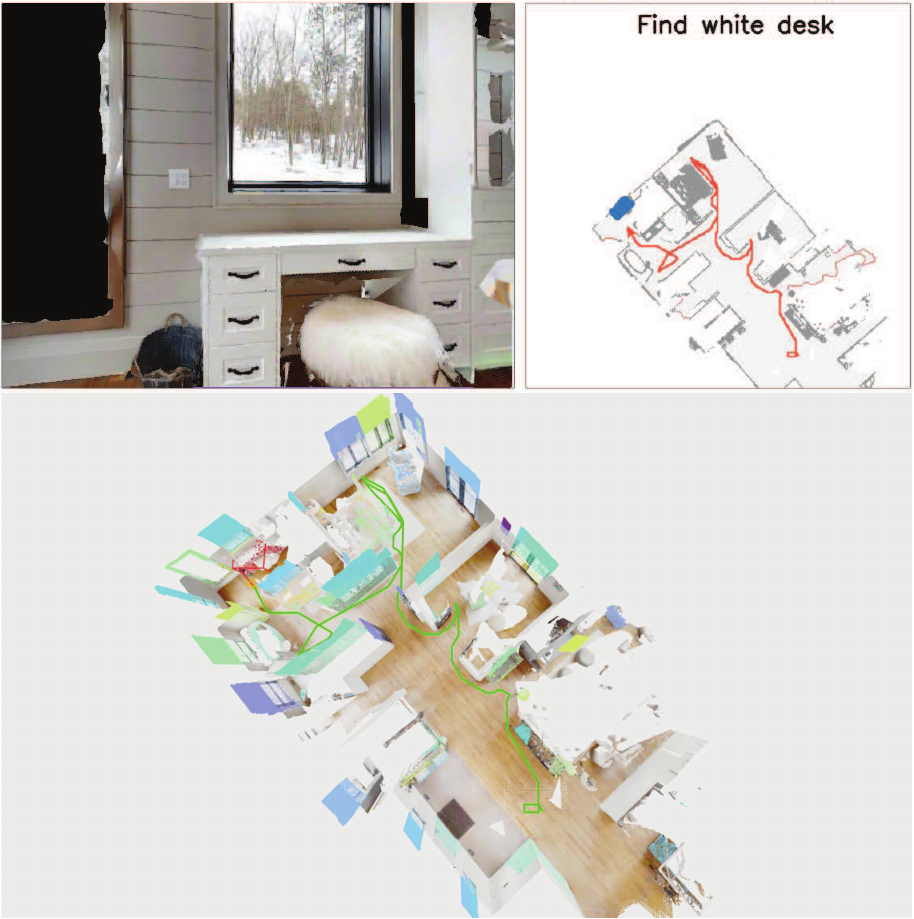}
}
\caption{The process of visual target navigation to find a desk located in front of the cabinet. The desk is described as a white desk with a window in front of it. Utilizing the CLIP model, the robot identifies a candidate desk at step 10; however, this desk is subsequently rejected by the game-theoretic vision-language model due to inconsistencies with the target description. After continuing the search through various iterations, the actual target desk is successfully identified at step 100.}
\label{fig:sim-vln}
\end{figure*}

To assess the navigation performance of our model on the language description evaluation datasets, a few baselines are considered: 

\begin{itemize}
    \item CLIP-based \cite{clip}: We use CLIP \cite{clip} to process the target language description as an open-vocabulary single target in the Object Goal Navigation task, then apply our exploration policy to find the target. CLIP selects the object with the highest similarity to the query's embedding.
    \item VLN-Generator\cite{openai2023gpt4}: this baseline uses only the Generator $\pi^{(1)}_G$ from our framework for language target identification. The main target is parsed from the language description and searched using the exploration policy. Then, the vision-language model (VLM) identifies which main target matches the description. We test the performance with GPT-4o and GPT-4o-mini \cite{openai2023gpt4}.
    \item VLN-Ranking\cite{ranking}: This approach ranks all candidates and selects the top candidate. It is the standard method used in previous work, as it can significantly reduce partial hallucinations and maintain better coherence. In this baseline, we sample five times from the VLN-Generator $\pi^{(1)}_G$ and rank the results to select the best outcome for language target identification.

 \end{itemize}

The quantitative results of our comparison study are reported in TABLE \ref{tab:vln_performance}. As reflected in the results, the CLIP-based method struggles to understand target language descriptions involving spatial relationships in the 3D environment, as it primarily models orderless content. Compared with CLIP, VLN-Generator performs better due to the strength of GPT models: GPT-4o-mini improves the success rate by 5 \%, while GPT-4o achieves nearly a 10\% increase. However, the SPL of GPT-4o-mini is similar to that of CLIP, as CLIP often grounds objects easily, regardless of correctness. The results for VLN-Ranking show that ranking significantly mitigates partial hallucination and improves coherence, leading to better performance compared to VLN-Generator. Our method (VLN-Game) outperforms all the baselines, regardless of whether GPT-4o or GPT-4o-mini is used. Notably, VLN-Game with GPT-4o-mini even surpasses the larger model, VLN-Generator with GPT-4o, demonstrating the significant advantage of game-theoretic vision-language models in language goal navigation tasks and highlighting the potential of smaller vision-language models in advancing robotic decision-making. We also observe that the improvement of our method (VLN-Game) compared to VLN-Ranking is more pronounced with GPT-4o-mini than with GPT-4o, indicating that our approach can significantly enhance the performance of smaller vision-language models. Fig. \ref{fig:sim-vln} shows the process of visual target navigation to find a desk located in front of the cabinet. The desk is described as a white desk with a window in front of it. There is more than one desk in the scene and the robot needs to identify which one satisfies our description.

\subsubsection{Failure Cases}
While our approach achieves high success rates across various tasks and environments, we have identified several failure cases and limitations that warrant future attention. Although the method is effective in finding open-vocabulary targets in both simulation and real-world environments, identifying these targets from multiple views remains challenging. We attribute this to two main causes:
First, since the CLIP model is used to locate the target based on the similarity between the object's CLIP features and the target's CLIP text features, the similarity scores can vary across different views of the same object. This inconsistency affects the accuracy of target identification during navigation. Second, the CLIP model has limitations in detecting all open-vocabulary targets in our tasks, which constrains the overall navigation performance of the robot. For tasks involving a wider range of open-vocabulary targets, the robot could benefit from leveraging more powerful contrastive language-image models.

\subsection{Navigation in Real World}

\begin{figure}[b]
    \centering
    \includegraphics[scale=0.42]{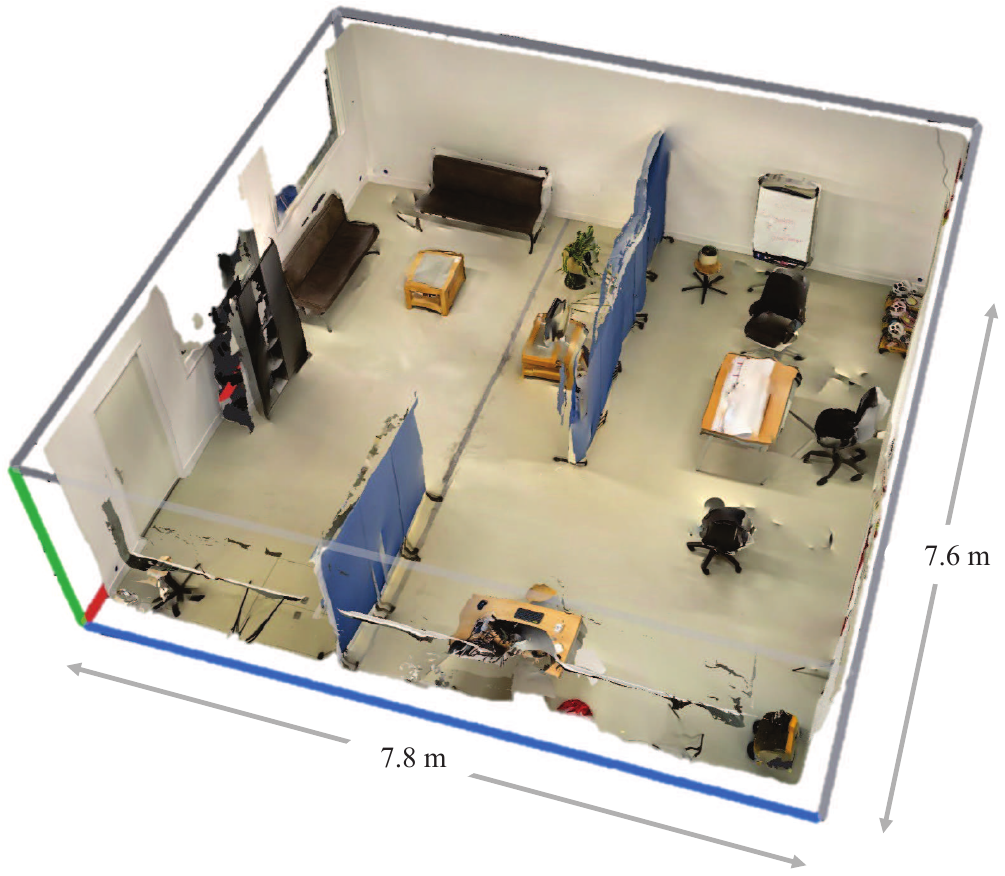}
    \caption{The scanned 3D reconstruction of the real scene.}
    \label{fig:realscan}
    \vspace{-0.5cm}
\end{figure}

\begin{figure*}[htbp]
\centering
\subfloat[Target: A brown sofa.]
{
    \includegraphics[width=8.75cm]{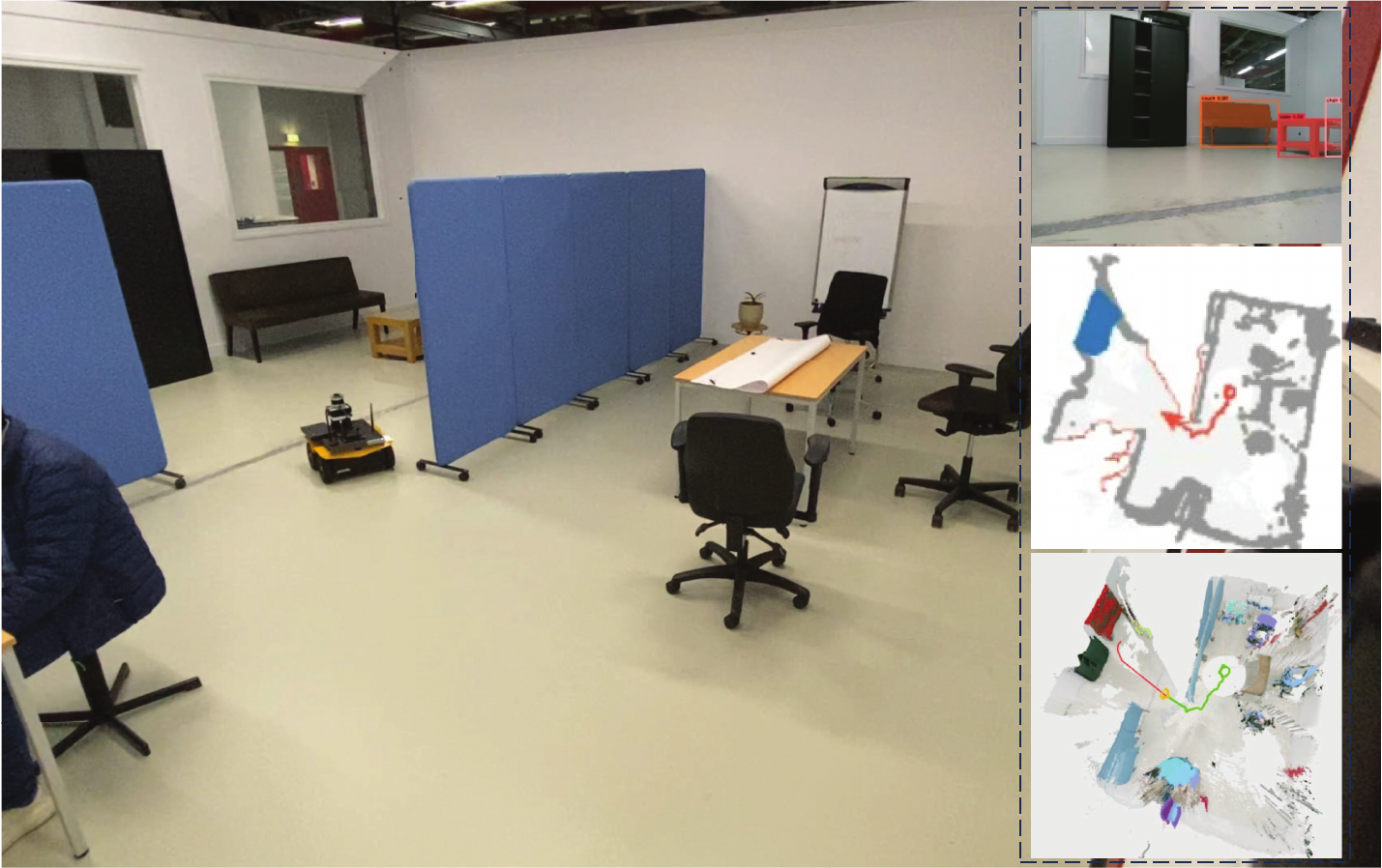}
}
\subfloat[Target: A brown sofa near a plant.]
{
    \includegraphics[width=8.75cm]{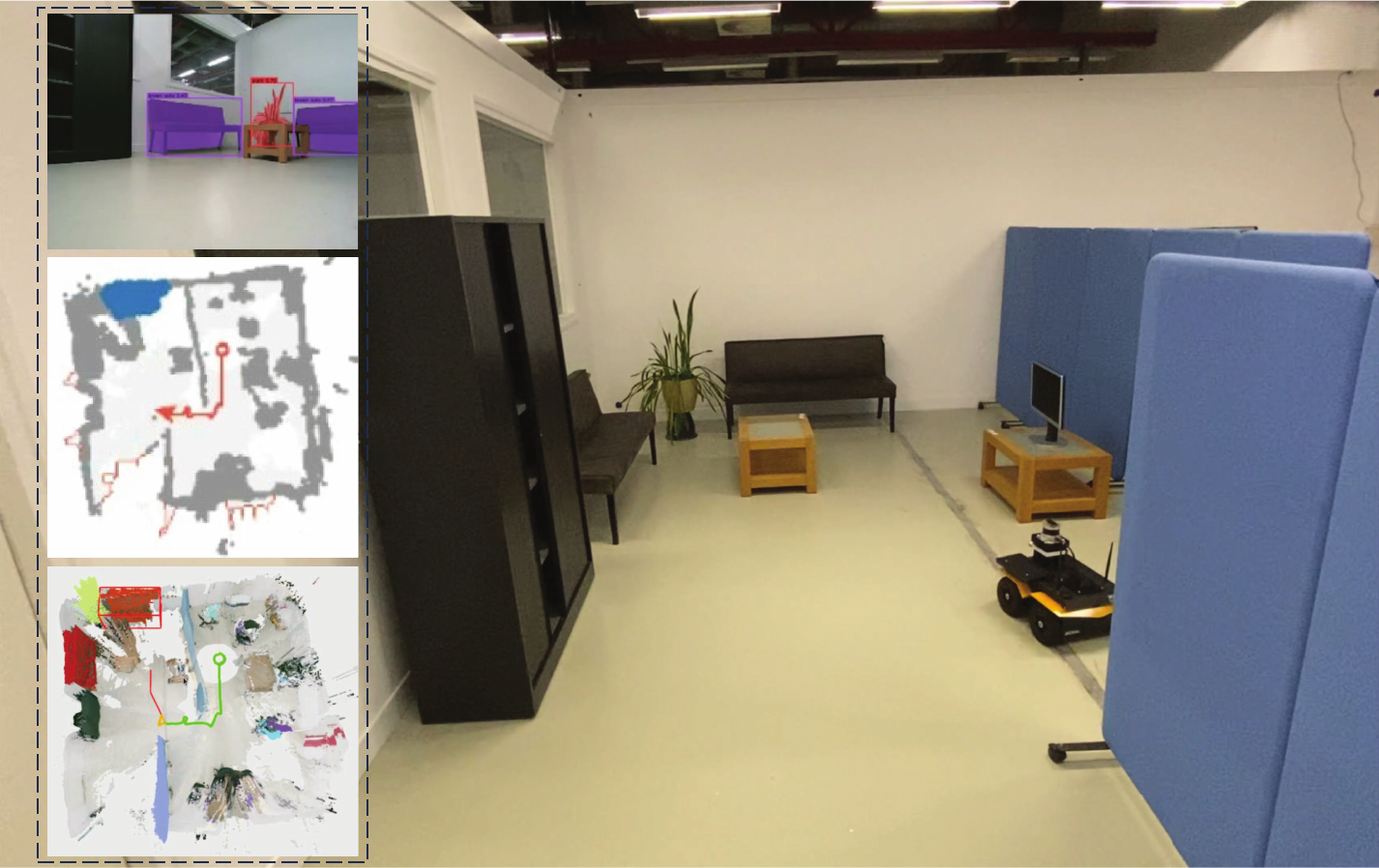}
}

\subfloat[Target: A black office chair.]
{
    \includegraphics[width=8.75cm]{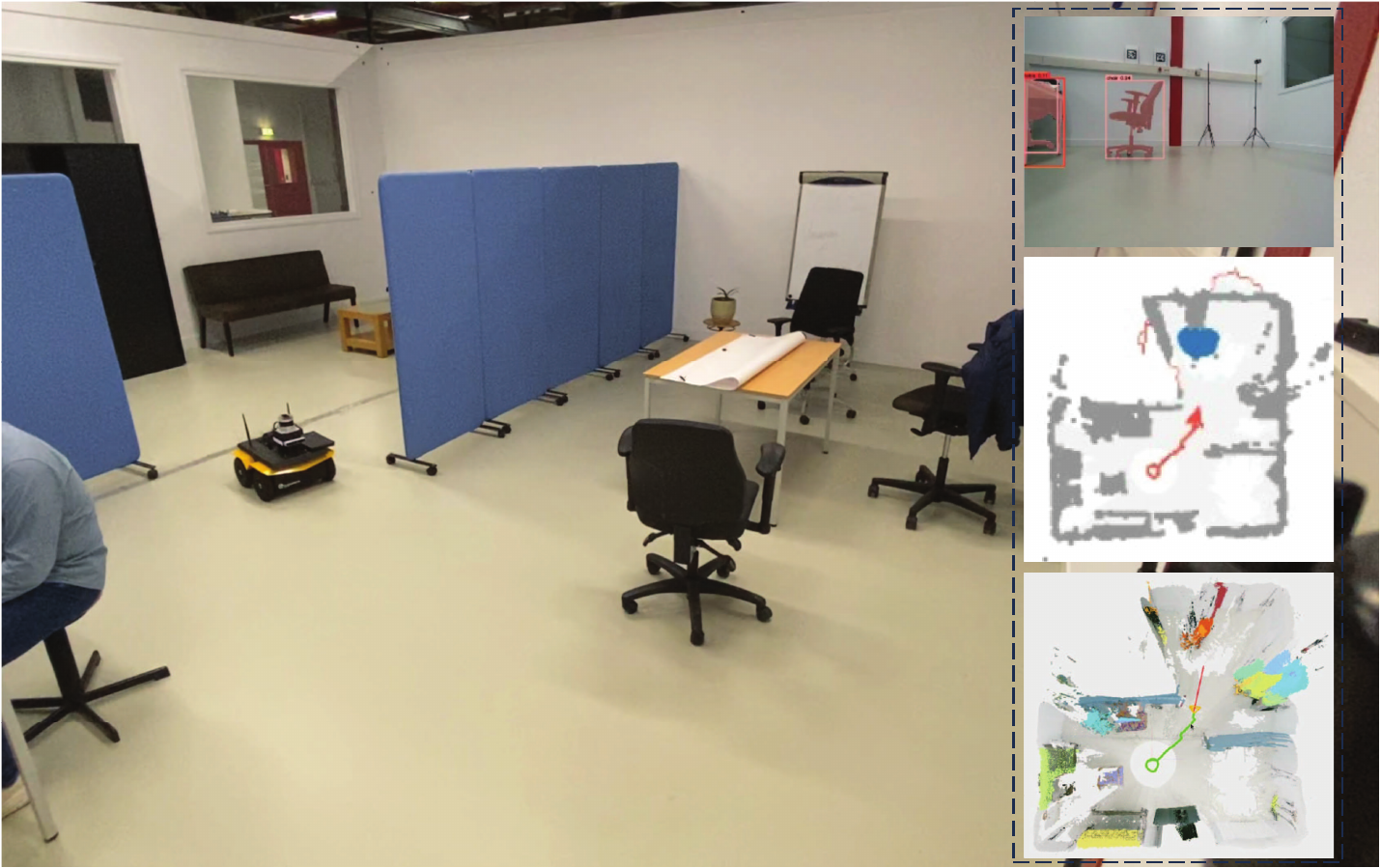}
}
\subfloat[Target: A black office chair between a whiteboard and a table.]
{
    \includegraphics[width=8.75cm]{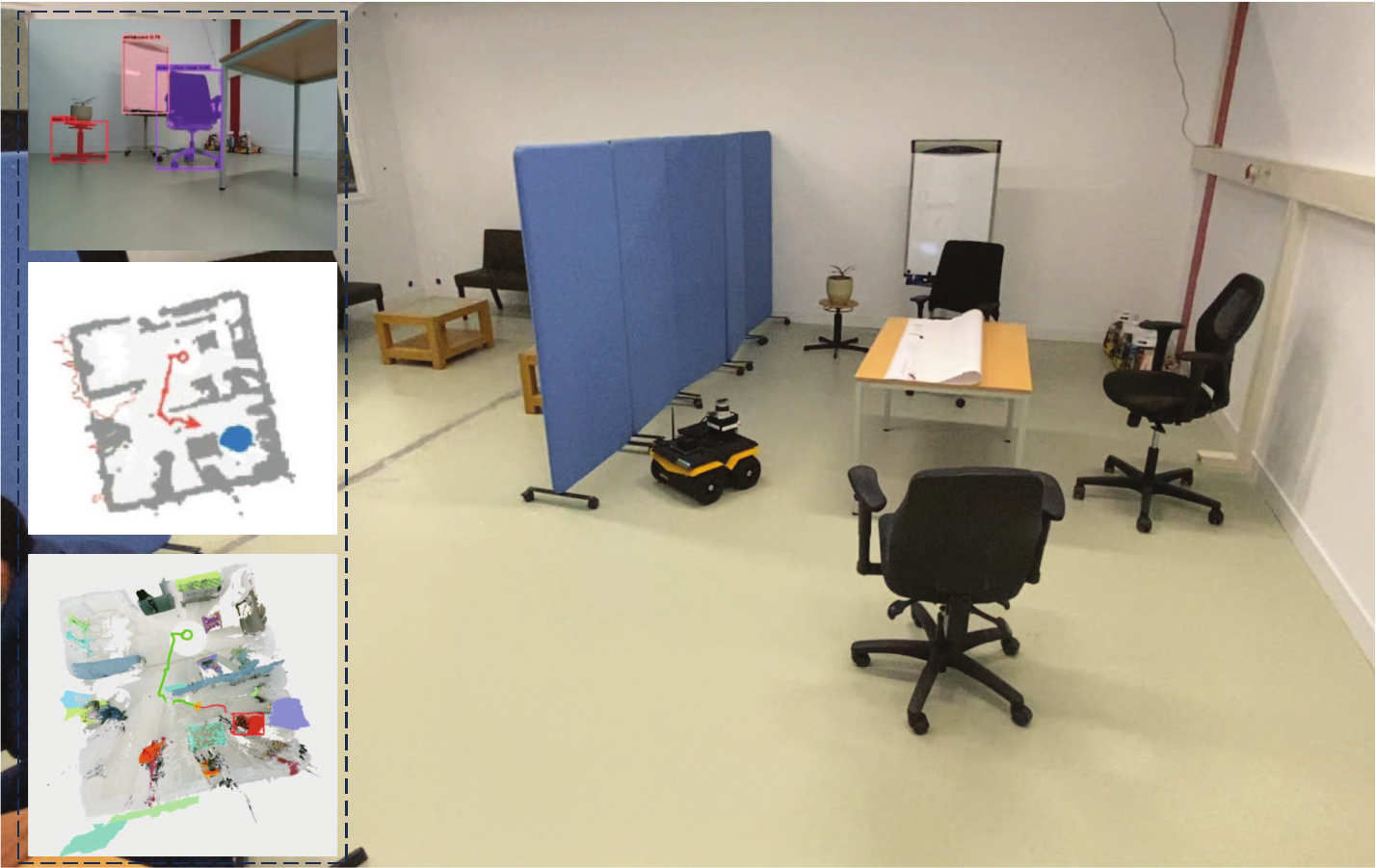}
}

\subfloat[Target: A sitting person]
{
    \includegraphics[width=8.75cm]{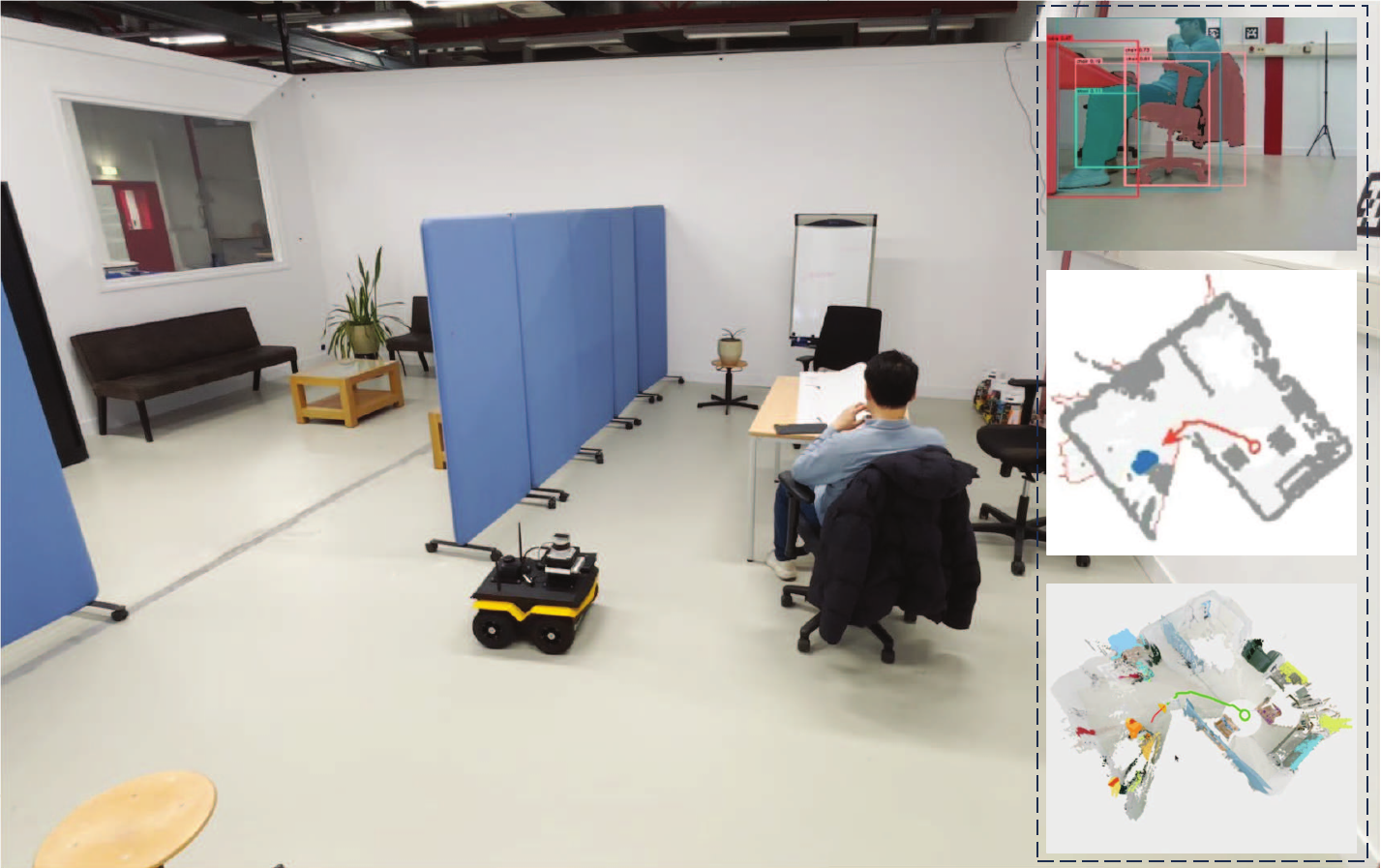}
}
\subfloat[Target: A person sitting on the sofa.]
{
    \includegraphics[width=8.75cm]{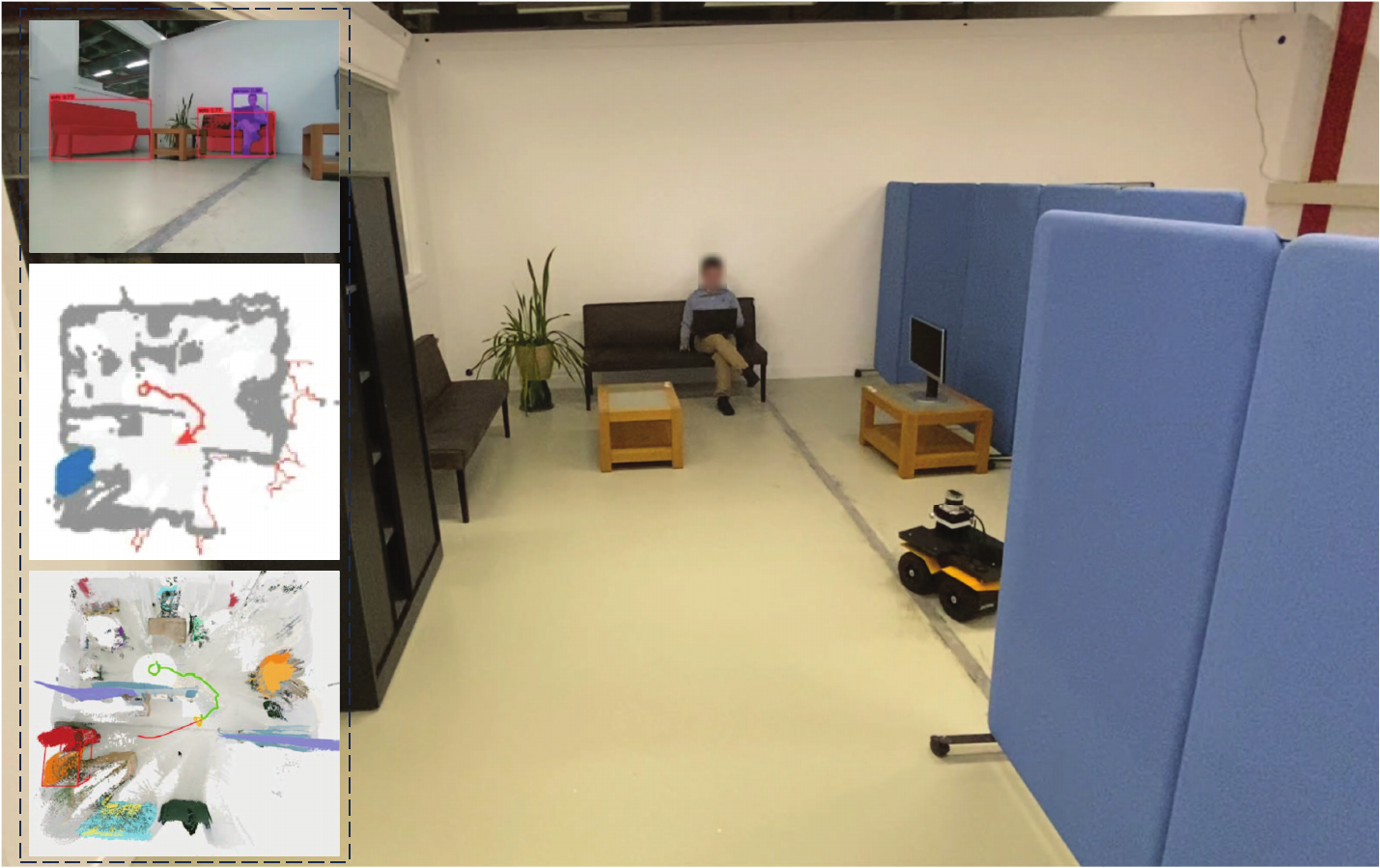}
}
\caption{Vision language navigation in a real office scene. This visualization demonstrates how the robot utilizes RGB-D images and localization to create a 3D object-centric map and an exploration map in a real office environment. The first-person perspective RGB Images, exploration maps, and the 3D object-centric maps are shown in each task.
The robot identifies targets by interpreting task-specific instructions through vision-language integration, enabling precise navigation and interaction with objects in the workspace. Specifically, (a), (c), and (e) illustrate the process of object goal navigation, where the robot aims to locate any one of the specified objects. In contrast, (b), (d), and (f) showcase language-descriptive goal navigation, where the robot filters out distractions and searches for a specific target that matches the spatial relationships described in the instructions.}
\label{fig:real}
\vspace{-0.4cm}
\end{figure*}

We deployed the agents in the real world and evaluated them on the Jackal Robot platform, which is equipped with a RealSense D455 RGB-D camera and an Ouster lidar, using ROS \cite{ros}. The evaluation was conducted in a complex and unknown office environment (Fig. \ref{fig:realscan}), with two navigation tasks: visual target navigation with the category's name and language description.

\begin{itemize}
    \item \textbf{Task 1: Object Goal Navigation} The robot is given the objects: \emph{A brown sofa, A black office chair, and a sitting person} and needs to explore the unknown environment and locate these targets.
    \item \textbf{Task 2: Language Goal Navigation} The robot is instructed to the language descriptive targets: \emph{A brown sofa near a plant, A black office chair between a whiteboard and a table, and a person sitting on the sofa.} There is only one object for each instruction in the environment. The robot must navigate through a complex office area while not mistakenly approaching other targets with different attire.
\end{itemize}

To bridge the gap between simulation and the real world, we ensure that all sensor data remains as consistent as possible with the simulation environment. First, we configure the RGB-D images to match the resolution and depth range settings used in the simulation. For accurate real-time localization and geometry map generation, we utilize Ouster lidar along with GMapping \cite{gmapping}. The RGB-D images, location data, and task instructions are then processed by our framework, which generates waypoints on the geometry map. As shown in Fig. \ref{fig:real}, a 3D object-centric map and an exploration map are created for each task, guiding the robot to explore the environment efficiently and locate the target. Our framework enables the robot to identify the correct objects based on language descriptions that include spatial relationships, helping it avoid similar-looking objects. For example, if the robot is tasked with finding the black office chair positioned between a whiteboard and a table, but encounters multiple chairs during exploration, it will check each chair until it identifies the one that precisely matches the description.
After determining target coordinates from frontiers or target objects during the exploration phase, the robot plans a trajectory using FMM\cite{fmm} to reach them. 

All the perception and navigation modules, including GroundingDINO \cite{dino}, MobileSAM \cite{SAM}, CLIP \cite{clip} and our navigation framework, are deployed on a computer equipped with an NVIDIA GeForce RTX 3060 GPU. The sensor, localization, and action modules, such as camera and GMapping \cite{gmapping}, run directly on the robot. Our approach can be implemented in the robot flexibly since only the image observation and the localization are necessary for our framework, and the output of our framework is the discrete action of the robot. The supplementary videos can be found via the following link: \href{https://sites.google.com/view/vln-game}{https://sites.google.com/view/vln-game}. 





\section{Conclusion}
\label{sec:conclusion}
We presented VLN-Game, a novel zero-shot modular framework that leverages game-theoretic vision language models to facilitate visual target navigation with an object's description by inferring the semantic similarity from the frontiers and identifying the attributes and spatial relationships of the observed target. By implementing experiments in the HM3D dataset, we demonstrate that our approach achieves state-of-the-art results on object-goal navigation and significantly improves the success rate and efficiency in language descriptive navigation tasks while just using a tiny model. We also validate the effectiveness in a real-world experiment, highlighting the practical applicability of our method. Our findings highlight the significant potential of game-theoretic vision-language models in assisting robotic navigation tasks by offering robust reasoning capabilities with lightweight architectures.
Future research could explore integrating more advanced vision-language models into this framework and extending its application to other challenging tasks, such as object delivery and lifelong navigation.



\bibliographystyle{ieeetr}
\bibliography{bib/library.bib}



	

\end{document}